%% file: main.tex
\documentclass[sigconf]{acmart}
\usepackage[a-2b]{pdfx}
\usepackage{balance}

\usepackage{amsmath}
\usepackage{fontawesome}
\usepackage{bm}
\usepackage{graphicx}
\usepackage{etoolbox}
\usepackage{enumitem}
\usepackage[noend]{algorithmic}
\usepackage{multirow}
\usepackage{amsthm}
\usepackage{xspace}
\usepackage{listings}
\usepackage{multirow}
\usepackage{diagbox}
\usepackage{balance}
\usepackage{float}
\usepackage[noend,lined,boxed,vlined,ruled,linesnumbered]
{algorithm2e}
\usepackage{caption, subcaption}

\usepackage[labelformat=simple]{subcaption}
\SetKwInOut{Input}{input}
\SetKwInOut{Output}{output}
\newcommand{\code}[1]{{\small \texttt{#1}}}
\newcommand{\sys}{\textsc{Table-GPT}\xspace}

\newcommand{\stitle}[1]{\vspace{1ex}\noindent{\bf #1}}

\newcounter{definition}
\newenvironment{definition}[1][]{\refstepcounter{definition}\par\smallskip\textsc{Definition~\thedefinition.\ #1}}{\smallskip}

\newcounter{example}
\newenvironment{example}[1][]{\refstepcounter{example}\par\smallskip\textsc{Example~\theexample.\ #1}}{\smallskip}

\newcounter{theorem}

\newcounter{proposition}

\newtoggle{fullversion}
\toggletrue{fullversion}
\newcommand{\revised}[1]{{\color{blue}#1}}

\begin{document}
\title{Table-GPT: Table-tuned GPT for Diverse  Table Tasks}

%\author{Anonymous author(s)}

% \author{Peng Li$^1$, Yeye He$^2$, Weiwei Cui$^2$, Song Ge$^2$ \\ Haidong Zhang$^2$, Shi Han$^2$, Dongmei Zhang$^2$, Surajit Chaudhuri$^2$}
% \affiliation{\vspace{0.1cm}
% \institution{$^1$ University of California, Los Angeles}
% \institution{$^2$ Microsoft Research}
% }

\author{Peng Li$^\dagger$, Yeye He$^\ddagger$, Dror Yashar, Weiwei Cui, Song Ge,  Haidong Zhang, \\ Danielle Rifinski Fainman, Dongmei Zhang, Surajit Chaudhuri}
\affiliation{\vspace{0.1cm}
%\institution{$^1$ University of California, Los Angeles}
\institution{Microsoft Corporation}
%\authornote{1: work done at Microsoft.}
}

% %
% % The "author" command and its associated commands are used to define the authors and their affiliations.
% \author{Peng Li}
% \authornote{Part of work done while at Microsoft.}
% \affiliation{%
%   \institution{Georgia Tech}
% }
% \email{pengli@gatech.edu}

% % \author{Yeye He, Cong Yan, Yue Wang, Surajit Chaudhuri}
% % \affiliation{%
% %   \institution{Microsoft Research}
% %   %\streetaddress{1 Th{\o}rv{\"a}ld Circle}
% %   %\city{Hekla}
% %   %\country{Iceland}
% % }
% % \email{{yeyehe, coyan, wanyue, surajitc}@microsoft.com}

% \author{Yeye He}
% \affiliation{%
%   \institution{Microsoft Research}
%   %\streetaddress{1 Th{\o}rv{\"a}ld Circle}
%   %\city{Hekla}
%   %\country{Iceland}
% }
% \email{yeyehe@microsoft.com}

% \author{Cong Yan}
% \affiliation{%
%   \institution{Microsoft Research}
%   %\streetaddress{1 Th{\o}rv{\"a}ld Circle}
%   %\city{Hekla}
%   %\country{Iceland}
% }
% \email{coyan@microsoft.com}

% \author{Yue Wang}
% \affiliation{%
%   \institution{Microsoft Research}
%   %\streetaddress{1 Th{\o}rv{\"a}ld Circle}
%   %\city{Hekla}
%   %\country{Iceland}
% }
% \email{wanyue@microsoft.com}

% \author{Surajit Chaudhuri}
% \affiliation{%
%   \institution{Microsoft Research}
%   %\streetaddress{1 Th{\o}rv{\"a}ld Circle}
%   %\city{Hekla}
%   %\country{Iceland}
% }
% \email{surajitc@microsoft.com}

%%
%% The abstract is a short summary of the work to be presented in the
%% article.
\begin{abstract}
%An emerging area of research in the natural language literature, called ``\emph{instruction-tuning}'', refers to the approach of using diverse \emph{(instruction, completion)} pairs to fine-tune pre-trained language models such as GPT-3, which has led to impressive models such as ChatGPT, that demonstrate remarkable capabilities to follow diverse human instructions and perform a wide range of novel tasks not seen by the model during training. %including table-manipulation tasks. 

Language models, such as  GPT-3 and ChatGPT, demonstrate remarkable abilities to follow diverse human instructions and perform a wide range of tasks. However, when probing language models using a range of basic table-understanding tasks, we observe that today's language models are still sub-optimal in many table-related tasks, likely because they are pre-trained predominantly on \emph{one-dimensional} natural-language texts, whereas relational tables are \emph{two-dimensional} objects.

%because unlike natural-language texts that are one-directional from left-to-right, tables are two-dimensional structures with both rows and columns, where reading in the vertical direction (for cells in the same columns) is equally if not more important for table understanding. We show that  today's language models do not sufficiently posses the ability of ``reading tables vertically'', because of the lack of such exercises in their pre-training or fine-tuning stages, which limits their performance on table-related tasks that are of interest to the database community.

%where permutations of tokens can lead to different meanings, tables are two-dimensional structures, where rows and columns that are largely permutation-invariant 

In this work, we propose a new  ``\emph{table-tuning}'' paradigm, where we continue to train/fine-tune language models like GPT-3.5 and ChatGPT, using diverse table-tasks synthesized from real tables as training data, with the goal of enhancing language models' ability to understand tables and perform table tasks. We show that our resulting \sys models demonstrate (1) better \emph{table-understanding} capabilities, by  consistently outperforming the vanilla GPT-3.5 and ChatGPT, on a wide-range of table tasks, including holdout unseen tasks, and (2) strong \emph{generalizability}, in its ability to respond to diverse human instructions to perform new table-tasks, in a manner similar to GPT-3.5 and ChatGPT. 

%where we start with pre-trained language models like the 175B GPT-3, and fine-tune using diverse ``\emph{(table-instruction, table-completion)}'' pairs, which we painstakingly collect from existing academic benchmarks and synthesize from real tables. We show that our resulting Table-GPT model functions like Chat-GPT, in its ability to follow diverse human instructions to perform tasks, while performing significantly better than the vanilla 175B GPT-3, on a wide-range of table tasks, including hold-out tasks completely unseen during table-tuning. 
\end{abstract}

\maketitle

\footnotetext{$\dagger$: Affiliation: Georgia Tech (pengli@gatech.edu), work done at Microsoft.}
\footnotetext{$\ddagger$: Correspondence: yeyehe@microsoft.com}

% %%% do not modify the following VLDB block %%
% %%% VLDB block start %%%
% \pagestyle{\vldbpagestyle}
% \begingroup\small\noindent\raggedright\textbf{PVLDB Reference Format:}\\
% \vldbauthors. \vldbtitle. PVLDB, \vldbvolume(\vldbissue): \vldbpages, \vldbyear.\\
% \href{https://doi.org/\vldbdoi}{doi:\vldbdoi}
% \endgroup
% \begingroup
% \renewcommand\thefootnote{}\footnote{\noindent
% This work is licensed under the Creative Commons BY-NC-ND 4.0 International License. Visit \url{https://creativecommons.org/licenses/by-nc-nd/4.0/} to view a copy of this license. For any use beyond those covered by this license, obtain permission by emailing \href{mailto:info@vldb.org}{info@vldb.org}. Copyright is held by the owner/author(s). Publication rights licensed to the VLDB Endowment. \\
% \raggedright Proceedings of the VLDB Endowment, Vol. \vldbvolume, No. \vldbissue\ %
% ISSN 2150-8097. \\
% \href{https://doi.org/\vldbdoi}{doi:\vldbdoi} \\
% }\addtocounter{footnote}{-1}\endgroup
% %%% VLDB block end %%%

% % %%% do not modify the following VLDB block %%
% %%% VLDB block start %%%
% \ifdefempty{\vldbavailabilityurl}{}{
% \vspace{.3cm}
% \begingroup\small\noindent\raggedright\textbf{Artifact Availability:}\\
% The data and/or other artifacts have been made available at \url{\vldbavailabilityurl}.
% \endgroup
% }
% % %%% VLDB block end %%%

\input{Introduction}

%\input{Related}
\input{Preliminary}
\input{CanLLMReadTable}
%\input{Method}
\input{FineTune}
\input{experiment}

\input{Conclusions}

%\clearpage

%\clearpage

%\balance

\bibliographystyle{ACM-Reference-Format}
\bibliography{TableGPT}

\iftoggle{fullversion}
{
    % removed for revision
    \revised{}
    \clearpage
    \appendix
    \input{apx-prompt}
}
{
}

\end{document}

%% file: Introduction.tex
\section{Introduction}
\label{sec:intro}

% \begin{figure}[t]
%     \centering    \includegraphics[width=1.0\columnwidth]{figures/chatgpt-analogy-2.png}
%     %\vspace{-5mm}
%     \caption{Table fine-tuning vs. Instruction fine-tuning}
%     \label{fig:all}
% \end{figure}
Large language models, such as GPT and LLaMa, have recently demonstrated impressive abilities in performing diverse natural-language tasks~\cite{llm-gpt-3, llm-palm-2, llm-llama, llm-palm}. In the database literature, a number of pioneering work, such as~\cite{stanford-prompt-engineer, cta-prompt-engineering, em-prompt-engineering, llm-vision}, %, datawrangling-prefix-tuning}, 
have also shown that by using  ``\emph{prompt engineering}'', to careful select the best instructions and few-shot examples for a particular task at hand, language models can be prompted to perform well on a number of table-tasks such as entity matching and data-imputation.

While prompt-engineering is a promising direction to enhance model performance, it requires task-specific tuning (e.g., task-specific labeled-data to test the performance of different instruction/example combinations)~\cite{llm-gpt-3, prompt-engineering, prompt-engineering-2}. We in this work
propose an orthogonal paradigm called ``\emph{table-tuning}'', where instead of modifying prompts, we modify the weights of the underlying  language models \emph{for once} (i.e., not task-specific), by continuing to train them using diverse table-tasks as training data, to improve their ability to understand tables.
We show that table-tuned \sys 
%by table-tuning GPT-3 and ChatGPT, respectively, the resulting models can better understand tables,  and consequently, 
consistently outperform the vanilla GPT-3.5 and ChatGPT on a wide-range of table tasks, including new and unseen table-tasks. % that are hold-out for testing only.
We note that our model-tuning approach is  \emph{complementary to} prompt-engineering, because carefully engineered prompts can continue to benefit both vanilla language-models and our table-tuned models.

%observe that today's pre-trained language models are not best-suited to read/understand tables, and therefore explore an orthogonal direction of ``\emph{table model-tuning}'', where we actually modify the weights of the underlying large language models, such as GPT-175B, \emph{for once only (i.e., not task-specific)}, so that the resulting model can better understand tables, and consequently, \emph{consistently perform better on a wide-range of table tasks}, including new unseen table-tasks that are hold-out for testing only.
%We show that our ``{table model-tuning}'' approach not only benefits complementary to ``prompt-engineering'' 

\textbf{Today's language models cannot  ``read tables'' reliably.}
%While the language models today are powerful and excel in natural-language understanding tasks, and are shown to work well even on table-tasks with a bit of prompt-engineering~\cite{cta-prompt-engineering, stanford-prompt-engineer, em-prompt-engineering},
While today's language models excel in natural-language tasks, we start by asking the question of whether these models are optimal for table-tasks, because after all, they are pre-trained predominantly on natural language texts, which are different from tables. 

More specifically, natural language texts are (1) \emph{one-directional}, (2) read \emph{left-to-right}, where (3) swapping two tokens will generally change the meaning of a sentence. %(e.g., ``the fox jumped over the lazy dog'' vs. ``the dog jumped over the lazy fox''). 
In contrast, relational tables are (1) \emph{two-dimensional} in nature with both rows and columns, (2) where reading \emph{top-to-bottom} in the vertical direction for values in the same column, is crucial in many table-tasks. Furthermore, unlike text, (3) tables are largely ``invariant'' to row and column permutations, where swapping two rows or columns do not generally change the semantic meaning of the table. 

\begin{figure}[t]
    \centering  \includegraphics[width=1\columnwidth]{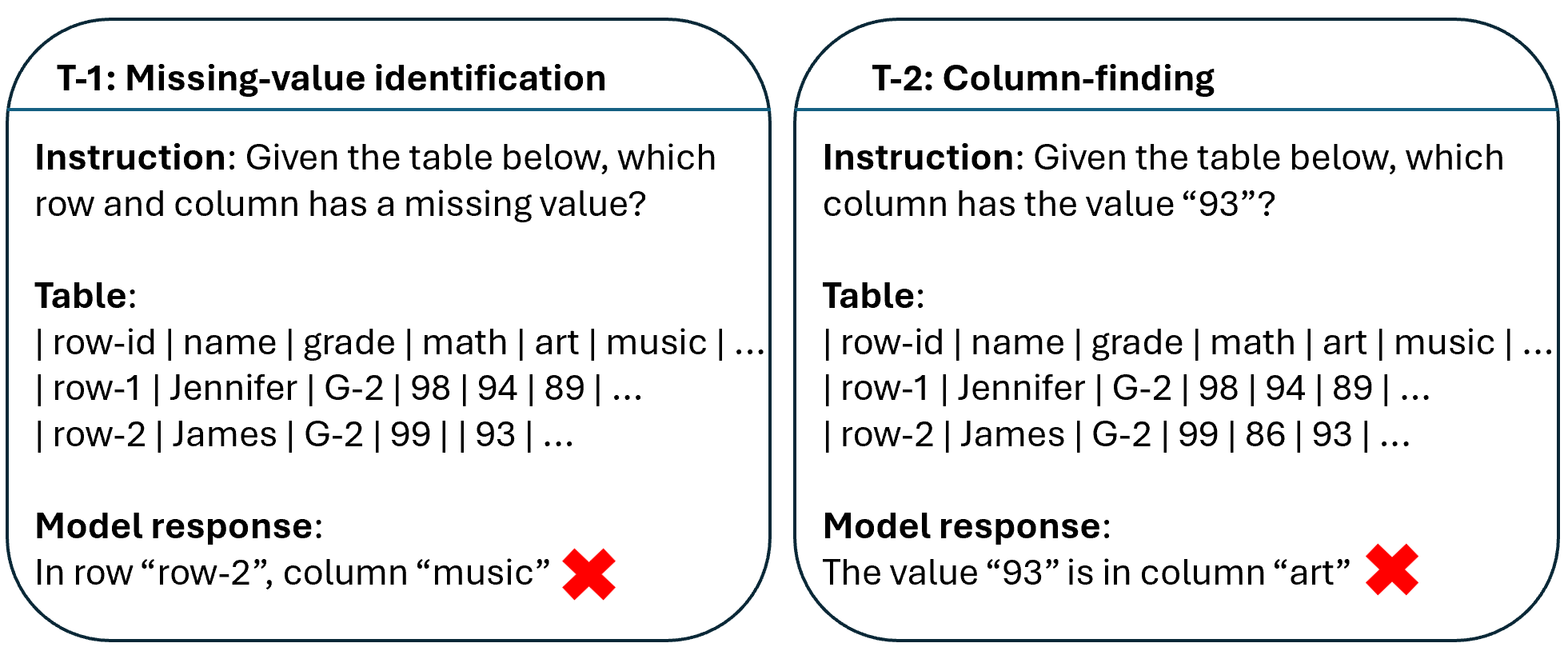}
    \vspace{-6mm}
    \caption{Two simple tests to probe language-models' basic ability  to read  and understand tables. (Left) T-1: Missing cells identification, which is to identify the column-header/row-id of a missing cell. (Right) T-2: Column-Finding, which is to identify the column-name of a given value. Even large models (e.g. 175B GPT-3.5) can frequently fail on such tests, with only 0.26 accuracy in one variant of the tests. % showing that models pre-trained on one-directional natural-language texts are not best-optimized for two-dimensional tables.
    }
    \vspace{-5mm}
    \label{fig:basic-tests}
\end{figure}

With this question in mind, we perform two simple tests to probe language models' ability to ``read'' tables and then answer basic questions, which we call (T-1) Missing-value-identification, and (T-2) Column-finding, as shown in Figure~\ref{fig:basic-tests}.   

In (T-1) Missing-value-identification, we show language models with a real table, presented in a markdown\footnote{Markdown table is a common format used by prior work to feed tables into language-models, and also a format that models like GPT will use  when it needs to respond with a table,  presumably because GPT-like models use GitHub data in its pre-training, where markdown-format tables are abundant.} or alternative format,  where we make sure that there is exactly one empty cell in the table. We then ask the model to identify the empty cell, by responding with the column-name and row-id of the empty cell, repeating for 1000 randomly sampled real tables. Despite the impressive ability of language-models like GPT-3.5 to perform diverse tasks, %complex table-tasks such as data imputation and entity-matching~\cite{em-prompt-engineering, stanford-prompt-engineer, cta-prompt-engineering}, 
we find that they fail on a surprisingly large fraction (up to 74\%) of such tests, often responding with incorrect column-headers or row-ids -- for instance, in the example shown in Figure~\ref{fig:basic-tests}, the model may answer that the column ``\code{music}'' has a missing cell, when the correct answer should be ``\code{art}''. 

In order to ensure that there is no ambiguity in what ``missing value'' or ``empty cell'' could mean to language models, we design a second and even simpler test, which we refer to as: (T-2) Column-finding, shown on the right of Figure~\ref{fig:basic-tests}. In this test, we present a language model with a real table, and ask it to find a specific cell-value that appears exactly once in the entire table (e.g., ``\code{93}'' in this example), and then respond with the column-name of the that value. We find that language models such as GPT-3.5 are prone to fail on such tests again (e.g., answering that ``\code{93}'' is in column ``\code{art}'' when the correct answer is ``\code{music}''), on over half of such tests. 

We believe these simple probes show that today's large language models, when pre-trained on large amounts of one-directional natural-language texts, are not best-suited to ``read'' two-dimensional tables, especially in the vertical direction, which however is crucial in performing many table-tasks. 

\begin{figure}[t]
\vspace{-3mm}
    \centering  \includegraphics[width=1\columnwidth]{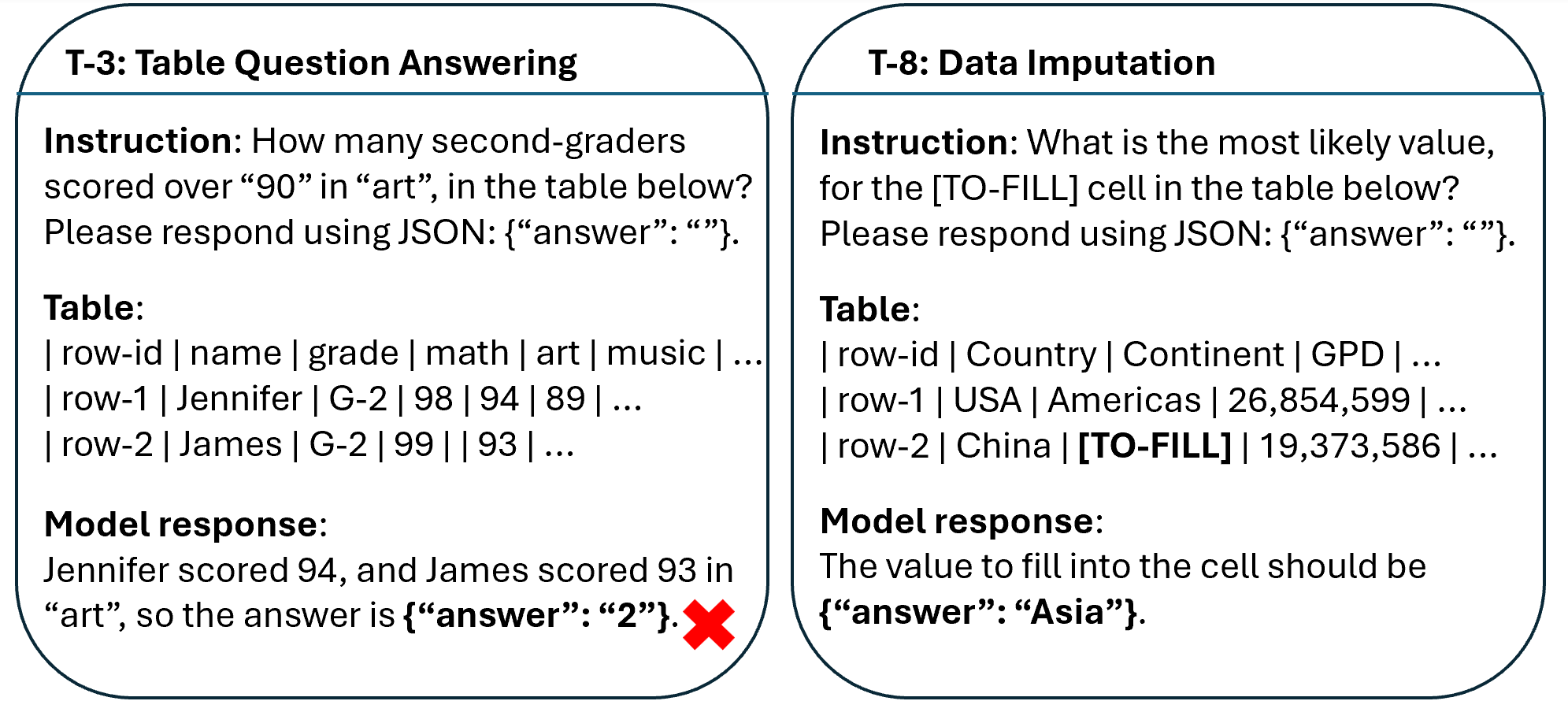}
    \vspace{-5mm}
    \caption{Example table-tasks, where the ability of language models to ``read'' tables vertically is important. (Left) T-3: Table Question-Answering. %, where answers are predicted for a natural-language question on an input table. 
    (Right) T-8: Data Imputation. %, where the value of a missing cell is predicted, based on its column/row context. 
    More tasks like these are shown in Table~\ref{tab:task-summary}.}
    \vspace{-5mm}
    \label{fig:more-tests}
\end{figure}

Consider, for example, the popular NLP task of (T-3) Table-QA~\cite{table-qa-tabfact, table-qa-wikitablequestions, table-qa-2}, %or NL-to-SQL~\cite{nl2sql-2, nl2sql-spider, nl2sql-wikisql}, 
where the task is to answer a natural-language question, based on the content of the table. The left side of Figure~\ref{fig:more-tests} shows such an example. Given the question ``\code{How many second-graders scored over 90 in art, in the table below?}'' Imagine that a model is not able to ``read'' tables correctly, it may reason that both ``\code{Jennifer}'' and ``\code{James}'' satisfy the condition (because it believes ``\code{93}'' is in the column ``\code{art}'', like shown in Figure~\ref{fig:basic-tests} (Right)), and may answer ``\code{2}'' instead of the correct ``\code{1}''.  We emphasize that the ability to read in the vertical direction (top-to-bottom for values in the same column) is similarly important in many other table-tasks, such as  data-imputation (shown on the right of Figure~\ref{fig:more-tests}), data-transformation, error-detection, NL-to-SQL, etc., like the list in Table~\ref{tab:task-summary} would show, which includes a diverse set of table-tasks considered in this work.

%We note that our observation that LM cannot reverse a sentecen

In addition, we find that large language models are sensitive to the order in which columns are presented in a table -- e.g., when we swap the order of two columns in a table, a model can change its response for a table-task, even when such a swap should not change the semantic meaning of the table, at least to humans.  This is presumably because language-models are pre-trained on text where the order of tokens matters (e.g., ``\code{Jennifer called you}'' vs. ``\code{you called Jennifer}''),  leading to sub-optimal behaviors on tables.

%and therefore become sensitive to the order of columns in a table, when such an order should largely not matter.  

We believe observations like these point to  opportunities for us to improve the underlying language model, by enhancing their ability to understand tables and  perform table-tasks.

\begin{figure*}[t]
    \vspace{-5mm}
    \centering  \includegraphics[width=2\columnwidth]{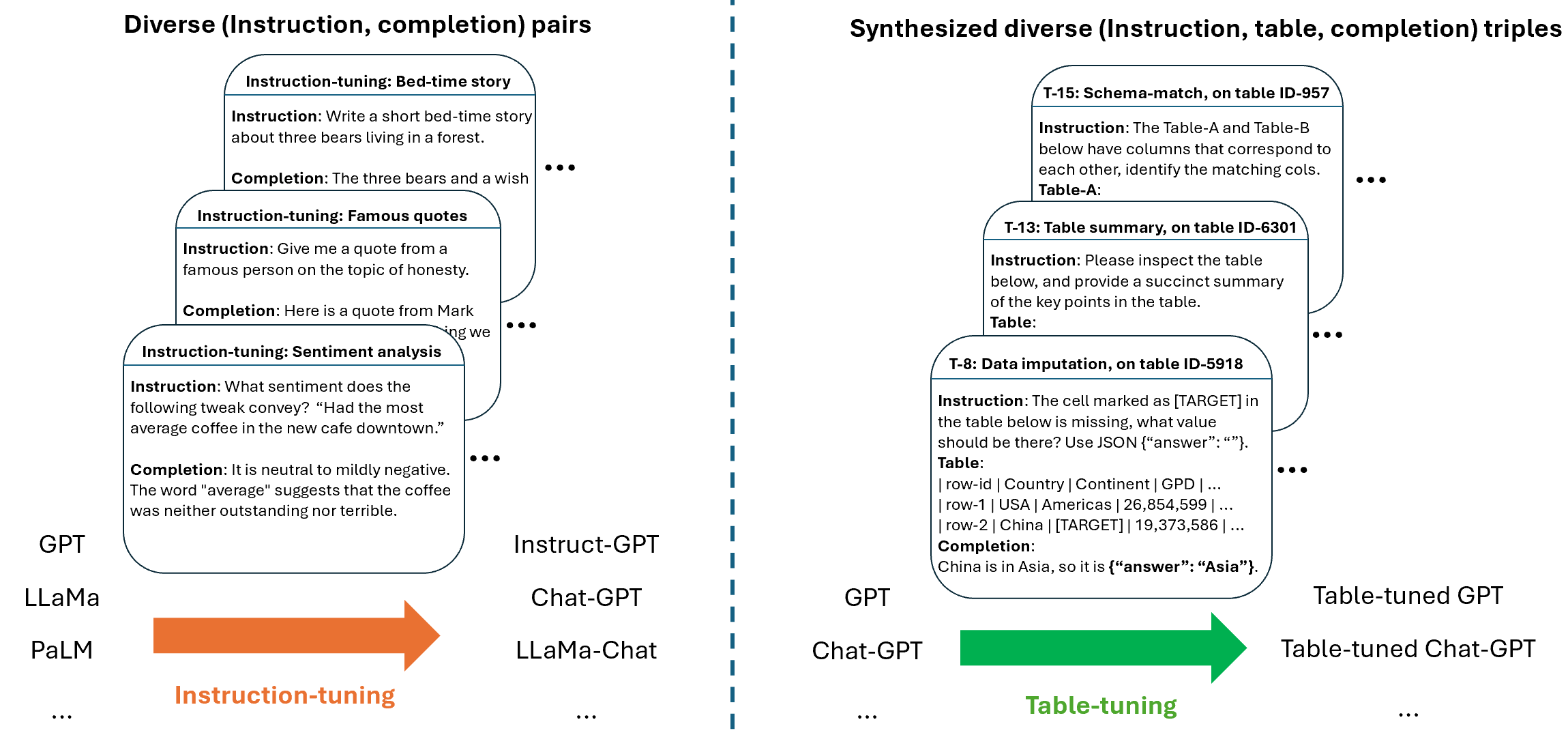}
    \vspace{-3mm}
    \caption{Instruction-tuning vs. Table-tuning. (Left) Instruction-tuning is a technique developed in the NLP community that continues to train language-models (e.g., GPT) for instruction-following capabilities (e.g., in ChatGPT). (Right) Table-tuning is an analogous approach we propose to train language-models to better understand table and perform table-tasks.}
    %\vspace{-5mm}
    \label{fig:instruction-tuning-vs-table-tuning}
\end{figure*}

\textbf{Instruction-tuning in NLP: train language-models to follow diverse human instructions.}
To change the behaviour of language models, successful attempts have been made in the NLP community, using a technique known as ``instruction-tuning'' in the literature~\cite{flan, instruct-gpt, self-instruct, t0, super-natural-instruction}. 

It was observed in the NLP community~\cite{flan, llm-gpt-3, instruct-gpt}, that earlier versions of pre-trained language models, such as GPT-3, is able to complete a sentence with the next likely token (e.g., ``\code{write a bed-time}'' $\rightarrow$ ``\code{story}''), but cannot reliable follow higher-level instructions from humans (e.g., ``\code{write a bed-time story for a 3 years-old, in 100 words}''), a behavior that is only demonstrated in later models such as ChatGPT. %, also known as zero-shot learning. Instead, users need to design few-shot prompts, with paired input/output examples (e.g., for sentiment analysis: ``\code{The food is amazing.}'' $\rightarrow$ ``\code{positive}''; ``\code{The service is quite slow.}'' $\rightarrow$ ``\code{negative}''), in order to demonstrate the task and reliably extract the desired outcome from the model.

Instruction-tuning was the key technique invented that continues to 
train GPT-like models into ChatGPT-like models, in a process shown on the left of Figure~\ref{fig:instruction-tuning-vs-table-tuning}. Diverse training data in the form of ``\code{(instruction, completion)}'' pairs are constructed, often manually annotated by human labellers~\cite{instruct-gpt}, e.g. (\code{``write a bed-time story'' $\rightarrow$ an-actual-story}), to continue train language-models on these explicit demonstrations of how to follow high-level human instructions, leading to well-known models such as ChatGPT/InstructGPT~\cite{instruct-gpt, chatgpt}, as well as their open-source counterparts like Stanford-Alpaca~\cite{stanford-alpaca} and LLaMa-chat~\cite{llm-llama}.

%to enhance language-models ability to follow diverse human instructions. Specifically large amounts of diverse ``\code{(instruction, completion)}'' training data pairs were collected and used to fine-tuning (or continue to train) GPT-like models, leading to highly well-known models such as InstructGPT~\cite{instruct-gpt}, ChatGPT~\cite{}, and their various open-source counterparts, all of which can reliably follow high-level human instructions, as Chat-GPT users can likely attest, even on novel tasks completely unseen during pre-training/fine-tuning time.

\textbf{Table-tuning: train language-models to understand tables.}
We believe that the research on instruction-tuning in NLP, which successfully enhances language-models ability to follow human instructions, holds lessons for us when we aim to enhance language-models ability to understand tables and perform table-tasks.

In this work, we propose a ``\emph{table-tuning}'' paradigm analogous to instruction-tuning, where we continue to train language-models, using diverse training data in the form of (\code{instruction, table, completion}), which we synthesize using large amounts of real tables. This process is illustrated on the right of Figure~\ref{fig:instruction-tuning-vs-table-tuning}.

%This leads us to ask the research question, of whether we can continue to train GPT, like how Chat-GPT is continuously trained from GPT using instruction-tuning, but instead of using (instruction, completion) pairs in instruction-tuning, we propose to use (table-instruction, completion) pairs generated from diverse table-related tasks, in a new paradigm we call ``table-tuning'', which is directly analogous to instruction-tuning proven to be highly effective. 

 Through extensive experiments, we show that ``table-tuning'' is a promising new direction, as our resulting \sys models are:
\begin{itemize}[noitemsep,topsep=0pt,leftmargin=*]
\item[] (1) \emph{\underline{Strong table models}}, which substantially outperform 175B GPT-3.5 and ChatGPT, on a wide range of seen and unseen table-tasks, as we summarize in Table~\ref{tab:task-summary} and Figure~\ref{fig:main-quality-chatgpt}; 
\item[] (2) \emph{\underline{Generalizable to new tasks}}, as they can respond well to novel and unseen table-tasks, similar to how Chat-GPT could generalize and respond to new and unseen NLP tasks, like shown in Figure~\ref{fig:unseen-new-tests}.
\end{itemize}

% (1) understand and respond to novel unseen (table or NLP) tasks, formulated in natural language questions, like shown in Figure~\ref{fig:basic-tests}; and (2) can understand tables better than vanilla-GPT, to not only answer the questions in Figure~\ref{fig:basic-tests} correctly (without seeing such tasks in the training process), but also performs substantially better than vanilla-GPT on a wide-range of table-tasks, like summarized in Table~\ref{tab:task-summary} and shown in Figure~\ref{fig:main-compare-gpt3}/Figure~\ref{fig:main-compare-cjatgpt}.

%%% [SIGMOD] to add back
%We perform an extensive number of over 1000 train and inference experiments before we arrive at successfully table fine-tuned models, which we believe are just first steps in this new direction.
%We report the lessons learned in the process, in the hope that our effort can serve as a springboard for new research in the direction of table-tuning, just like how instruction-tuning has become a fruitful line of research in NLP~\cite{self-alignment, self-instruct, instruct-gpt, chatgpt, flan, super-natural-instruction}. 

\textbf{Contributions.} We make the following contributions:

\begin{itemize}[noitemsep,topsep=0pt,leftmargin=*]
\item We propose a new ``table-tuning'' paradigm  to continue to train language models, specifically designed to enhance language-models' ability to  perform table-tasks, using diverse table-tasks synthesized from large amounts of real tables, in a ``synthesis-then-augment'' process. %We show that the resulting \sys is substantially better in performing table tasks than the vanilla GPT-3.5 or ChatGPT, including on \textit{novel unseen tasks}, showing the generality of \sys.
%\item We synthesize diverse table-tasks using large amounts of real tables (2M web tables and 10K database tables), which we show is crucial to ensure model generalizability and avoid over-fitting. %in our ablation analysis.
\item We develop task-level, table-level, instruction-level, and completion-level data augmentation techniques for table-tuning, which we show are crucial to avoid over-fitting and ensure the generality of \sys.
\item We show that \sys not only excels on table-tasks in both zero-shot and few-shot settings out of box, but can also serve as a ``table foundation model'' and used as a better starting point than vanilla GPT,  for down-stream single-task optimizations such as task-specific fine-tuning and prompt-engineering.
%\item We perform an extensive number of over 1000 train and inference experiments before we arrive at successfully table fine-tuned models, with substantial costs. We report the lessons learned in the process, and will share out code/data to facilitate new research in the promising direction of ``table-tuning''. %just like how the NLP community benefits tremendously from each other's efforts.
\end{itemize}
%\yeye{need a figure to show its generalizability to new unseen tasks? e.g., one cta with variable num of classes, and one extract tables from nl?}

%\yeye{add an figure example to show nl-to-sql requires reading tables correctly }

%% file: Preliminary.tex
%\vspace{-2mm}
\section{Preliminaries}
\label{sec:preliminary}

We will start with a review of language models, and then the use of language models in table-tasks.

%\subsection{Encoder vs. Decoder style language models}
\subsection{Language models}
There are two popular styles of language models today, known as the decoder and encoder-style, both derived from the original transformer architecture~\cite{transformer}.

\textbf{Encoder-style language models.}
One class of popular language models, including the well-known BERT~\cite{llm-bert} and RoBERTa~\cite{llm-roberta}, use only encoders from the transformer, and are pre-trained on large amounts of texts to effectively represent the semantics of texts using embedding vectors. 

\underline{Down-stream tasks: Task-specific fine-tuning.}
To use encoder-style models like  BERT for downstream tasks, \emph{task-specific fine-tuning} is generally employed~\cite{finetune-bert, finetune-bert-2}, which continues to fine-tune (or train) BERT-like models for a given task, using task-specific labeled data.  For example, suppose the downstream task is  sentiment analysis of Yelp restaurant reviews, then labels in the form of (``\code{The food is amazing}'', ``\code{positive}''), (``\code{The service is slow}'', ``\code{negative}''), are needed to fine-tune BERT-like models for the desired outcome~\cite{bert-study, llm-bert}.

Crucially, when the target input data or the desired output changes,  the labeling effort often needs to repeat for the best performance. For example, if the input data for sentiment analysis changes to IMDB reviews, or if the output needs to include a classification of ``\code{cuisine-type}'' for restaurant reviews. %(When a task is encountered, such as question-answering or machine-translation instead of sentiment-analysis,  a new set of labels is again necessary. 
While encoder-style language-models are strong models, the need to fine-tune with task-specific labelled data limits its ability to generalize to new unseen tasks~\cite{llm-bert, bert-study, adapt-llm, llm-roberta}.

%models can be used in specific downstream tasks (e.g., ), To perform a   These models are strong in representation, but are not generative, and  would typically require ``task-specific training''  for different downstream tasks.

\textbf{Decoder-style ``generative'' language models.}
Another class of decoder-only language models,  such as GPT~\cite{llm-gpt-3} and LLaMa~\cite{llm-llama}, are generative in nature, %that use only decoders from the transformer-architecture, 
and are shown to excel in generalizing to new downstream tasks \emph{without} task-specific fine-tuning~\cite{llm-gpt-3}.
%are gaining popularity. Crucially, unlike BERT-like encoder-style models, these decoder-style models are generative in nature, that can easily adapt to new tasks and datasets, with only a couple examples. % with few-shot learning.

%\yeye{ prefix -> next-token training objective}

\underline{Generalize to new tasks: zero-shot and few-shot learning.}
It was shown in the NLP literature that the decoder-style models (e.g., GPT and LLaMa), especially after instruction-tuning~\cite{flan, self-instruct, super-natural-instruction, t0, tulu, self-alignment, instruct-gpt, lima} (e.g., ChatGPT/InstructGPT~\cite{instruct-gpt, chatgpt} and Stanford Alpaca~\cite{stanford-alpaca}), can adapt to new tasks easily, using just natural-language instructions (e.g., ``\code{classify the sentiments in the following reviews}''), and optionally a few examples.
Such an approach can adapt to new datasets (e.g., IMDB vs. Yelp reviews) and new tasks (sentiment-analysis vs. machine-translation), without fine-tuning on labelled data for each specific task, making the decoder-style models more general and versatile. Figure~\ref{fig:models} shows the benefit of ``instruction-tuning'' in model generalizability, depicted pictorially on the y-axis.

\subsection{Language models for table tasks}
Pioneering work in the database literature have employed language models in various ways to perform table-related tasks.

\textbf{Encoder-style language models for table tasks}. There is a long and fruitful line of research (e.g., TURL~\cite{turl}, TaBERT~\cite{tabert}, Ditto~\cite{em-ditto} and Doduo~\cite{doduo}), where table-models are trained based on encoder-style BERT-like models, which are shown to perform well on various table tasks. 

\begin{figure}[t]
\vspace{-3mm}
    \centering  \includegraphics[width=1\columnwidth]{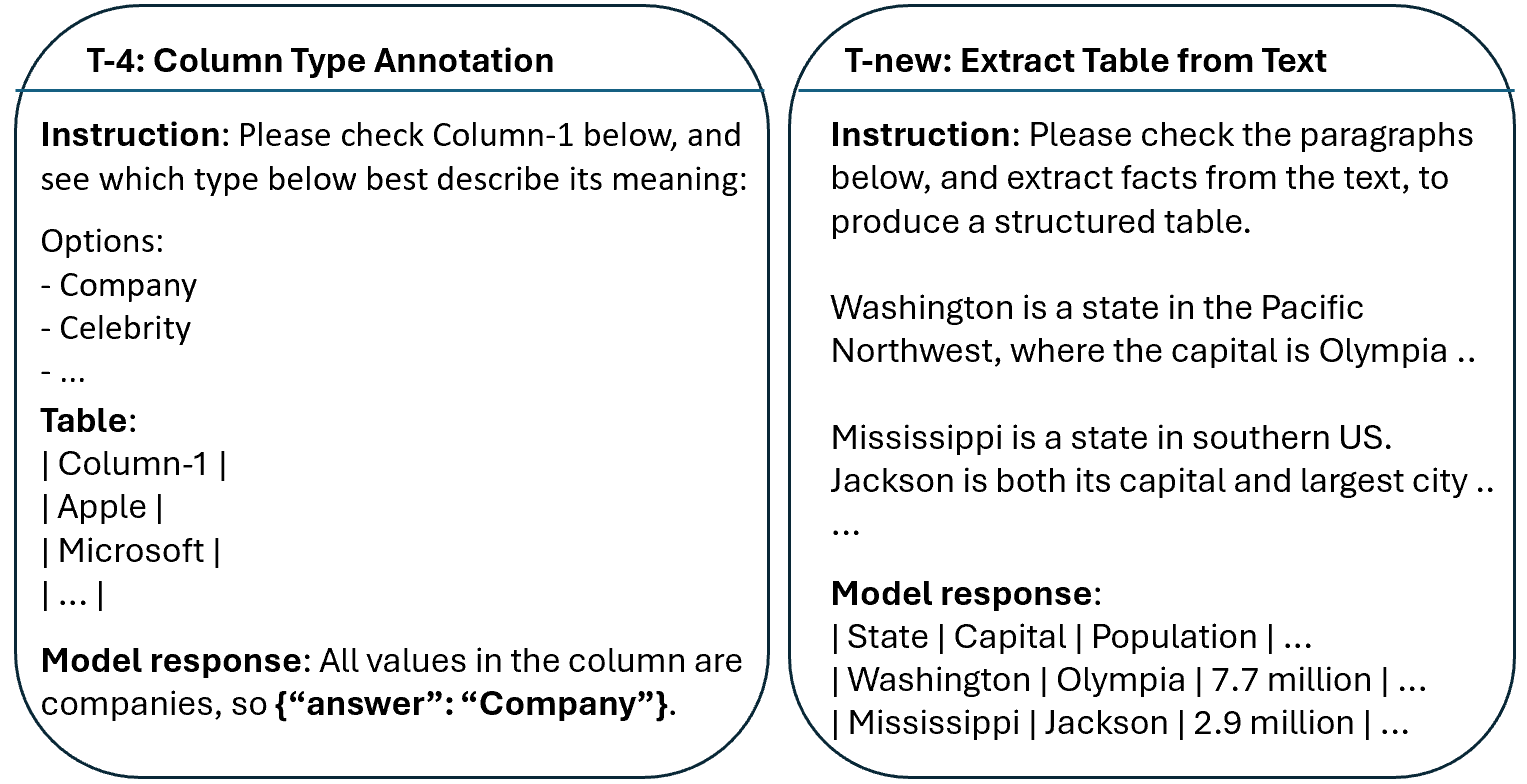}
    \vspace{-5mm}
    \caption{Table-models should ideally ``generalize'' to new datasets and new tasks. (Left) Column type annotation (CTA): while this is a common table-task, the list of target-types to choose from can vary from dataset to dataset (e.g., 78 types in~\cite{cta-sherlock}, and 107 in~\cite{turl}). Making table-models to ``generalize'' to new CTA dataset without needing to retrain, is useful. (Right) Text-to-Table: a general table-model should be as general-purpose as models like ChatGPT, in following instructions to perform novel unseen table-tasks, such as ``extracting tables from text'' in the example. 
    }
    %\vspace{-5mm}
    \label{fig:unseen-new-tests}
\end{figure}

However, like their BERT-like base models, to generalize to a new dataset or a new task, these encoder-style table-models generally require fine-tuning with labeled data. As a concrete example, for the table-task of ``column-type-annotation''~\cite{doduo, turl}, in order to move from one dataset with 78 semantic types~\cite{cta-sherlock}, to another dataset with 107 semantic types~\cite{turl}, new labeled data have to be obtained, so that the models can be fine-tuned to generate the new output with 107 classes~\cite{turl}.    In contrast, being able to adapt to new datasets and tasks \emph{without} task-specific fine-tuning, is a key goal that we want to achieve in this work, like illustrated in Figure~\ref{fig:unseen-new-tests}.

%The research literature is full of fruitful lines of explorations over the past decades, addressing a wide array of table-related tasks, such as 
%schema matching~\cite{schema-mapping-survey, schema-matching-valentine, schema-matching-cupid}, 
%entity matching~\cite{em-book, em-deepmatcher, em-ditto}, 
%data transformation~\cite{data-transform-tde, data-transform-flashfill, data-transformation-wrangler}, 
%data cleaning~\cite{error-detection-survey, data-cleaning-survey}, 
%list extraction~\cite{list-extraction-google, list-extraction-tegra, list-extraction-2}, 
%column type annotation~\cite{cta-autotype, cta-sherlock, turl}, 
%data imputation~\cite{turl, imputation-1, imputation-2},
%table-QA~\cite{table-qa-wikitablequestions, table-qa-tabfact, table-qa-2},
%NL-2-SQL~\cite{nl2sql-spider, nl2sql-wikisql, nl2sql-2}, 
%table augmentation~\cite{table-augmentation-2, table-augmentation-infogather}, 
%and table summary~\cite{table-summary-google, table-summary-2, table-summary-3}. 

\textbf{Decoder-style  language models for table tasks}. With the success of decoder-style  language models such as GPT-3 and ChatGPT, which are shown to perform tasks out-of-the-box with  instructions only, pioneering research in the database field develop ``\emph{prompt-engineering}'' techniques for table-tasks~\cite{stanford-prompt-engineer, em-prompt-engineering, cta-prompt-engineering}, which carefully selects instructions and examples in the prompt, such that vanilla language models can perform well on table-related tasks. 
%to optimize model performance on table-tasks. Authors in the pioneering work such as~\cite{stanford-prompt-engineer, em-prompt-engineering, cta-prompt-engineering} show that with careful prompting,  vanilla language models such as GPT-3 can perform well on a variety of table-related tasks. 

\underline{Table-tuning for table-tasks.}
In contrast to prompt-engineering that optimizes prompts, our proposed ``table-tuning'' explores the orthogonal direction, where we continue to train  the underlying language models, for once only (not task-specific), so that the resulting model perform better on a range of table-tasks. This is complementary to prompt-engineering, because carefully-engineered instructions and examples can continue to benefit both the vanilla GPT as well as our \sys, as we will show in our experiments. 

%In additional to model fine-tuning, ``\emph{prompt-engineering}'' is an orthogonal class of techniques that improve language-model capabilities on downstream tasks but without changing model weights, by carefully crafting instructions and examples selected in the prompt that is given to language models. 

\begin{figure}[t]
    \centering  \includegraphics[width=1\columnwidth]{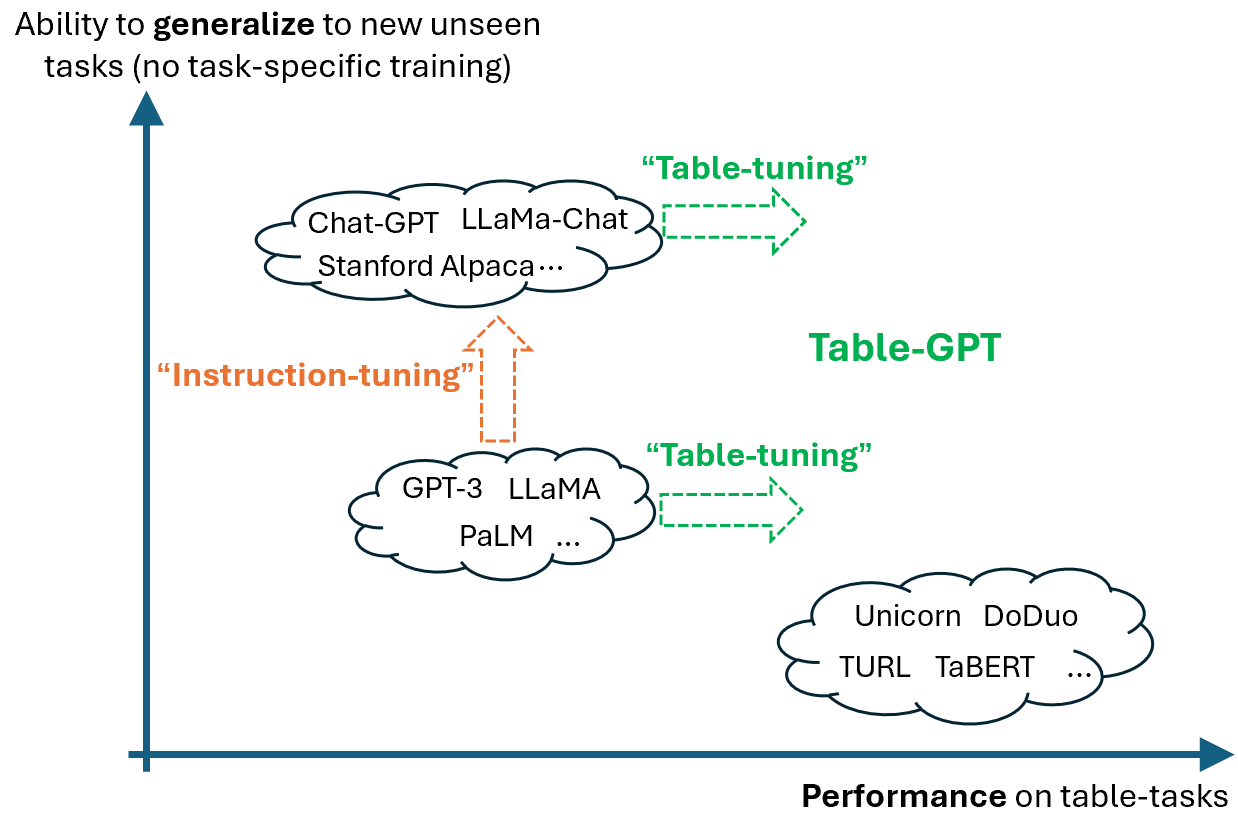}
    %\vspace{-5mm}
    \caption{Instruction-tuning vs. Table-tuning. Instruction-tuning improves model ``generalizability'', to follow diverse human-instructions to perform new and unseen tasks (x-axis), whereas our proposed table-tuning is analogous in spirit but aims to improve model ability to understand tables and perform  table-tasks (y-axis). %The proposed ``table-tuning'' is analogous to ``instruction-tuning'' from the NLP literature, where our hope is to open a new avenue that can train large language models to better understand table-tasks, while skill keeping models' generalizability, similar to what instruction-tuning achieves.
    }
    \vspace{-5mm}
    \label{fig:models}
\end{figure}

Figure~\ref{fig:models} shows the process of table-tuning, which is analogous to instruction-tuning, but unlike instruction-tuning that improves model generalizability to follow human instructions (y-axis), we focus on improving underlying models ability to understand tables and perform table-tasks (x-axis).
Crucially, as we will show, our table-tuned models remain to be general and capable of following human-instructions to perform table-tasks (without task-specific fine-tuning), just like the underlying GPT-3 and ChatGPT models. In other words, in \sys we aim to get the ``best of both worlds'', with both  generalizability, and good table-task performance.

%Via instruction-tuning, capable models like Chat-GPT and Stanford Alpaca are produced from pre-trained models like GPT-3 and LLaMa, whose ability to follow instructions and generalize to new tasks is enhanced (x-axis). Our proposed ``table-tuning'' would take a similar approach and aim to retain the resulting model's ability to follow human instruction to perform to novel and unseen tasks (x-axis), to achieve (Goal-\#1) above, but at the same time can also better understand tables and perform  better on table-tasks (y-axis), or the (Goal-\#2) above.

%We note that in the literature, there are also a number of strong table-representation models such as TURL~\cite{turl}, TaBERT~\cite{tabert}, etc., which are shown to achieve strong performance on table-tasks, or Goal-\#2, like shown in the figure. However, these existing methods either require task-specific training (thus requiring new labeled data for each task), or focus on certain categories of table-tasks (e.g., table-matching for Unicorn~\cite{unicorn} and table-annotation for DoDuo~\cite{doduo}), which could not directly follow instructions to perform novel and unseen table-tasks, like how Chat-GPT would behave. In other words, they are still lacking in Goal-\#1 above.

%In this work, we would like to get ``the best of both worlds'', to achieve both Goal-\#1 and Goal-\#2 above at the same time.  In effect, we would like to build a ``Chat-GPT for tables'', which has Chat-GPT-like generalizability to new tasks and instructions, and table-model's ability to understand tables.

%% file: CanLLMReadTable.tex
%\vspace{-2mm}
\section{Can Language Models ``read'' tables?}
\label{sec:read_tables}
Since language models like GPT are pre-trained predominantly on natural language text, we start by asking a basic question of whether language models can reliable read and understand relational tables, which are different from text in many ways, as we discuss below.

\underline{One-dimensional (text) vs. two-dimensional (tables)}. 
Language models trained mostly on natural language text (e.g, books and web pages) and programming code (e.g., GitHub), both of which that are \emph{one-directional} that is meant to be read \emph{left-to-right}, toke-by-token, in a sequential manner. 

In contrast, relational tables are  \emph{two-dimensional} with rows and columns, where reading \emph{top-to-bottom} vertically, for column-headers and other values in the same column (which may be far away when a table is serialized), is crucial for many table-tasks.

Consider the task of Data-Imputation~\cite{imputation-1, imputation-2}  (T-8 in Table~\ref{tab:task-summary}), which is to infer a missing value in a table cell, like shown in the example of Figure~\ref{fig:more-tests} (Right). At least for humans, it is natural to look vertically in the horizontal direction, to see the column-header (``\code{continent}'' in this case), as well as other values in the same column (e.g., ``\code{Americas}''), before one can make a guess for the missing value. 

Similarly, for the task of Error-Detection~\cite{error-detection-survey}  (T-9 in Table~\ref{tab:task-summary})  it is also necessary to look at the column-header and other values in the same column, to understand the semantics  of the column, before one can determine if a cell is erroneous.

Even for table-tasks that may be a bit removed, such as  Table Question-Answering~\cite{table-qa-2, table-qa-wikitablequestions} (T-3 in Table~\ref{tab:task-summary}), which is traditionally an NLP problem -- examples like in Figure~\ref{fig:more-tests} (Left) would show that, in order to answer a question correctly on a table, reading vertically in a column (e.g., for values in the \code{art}) is similarly important.

%\yeye{peng to check the numbers below.}

\begin{figure}[t]
    \centering  \includegraphics[width=1\columnwidth]{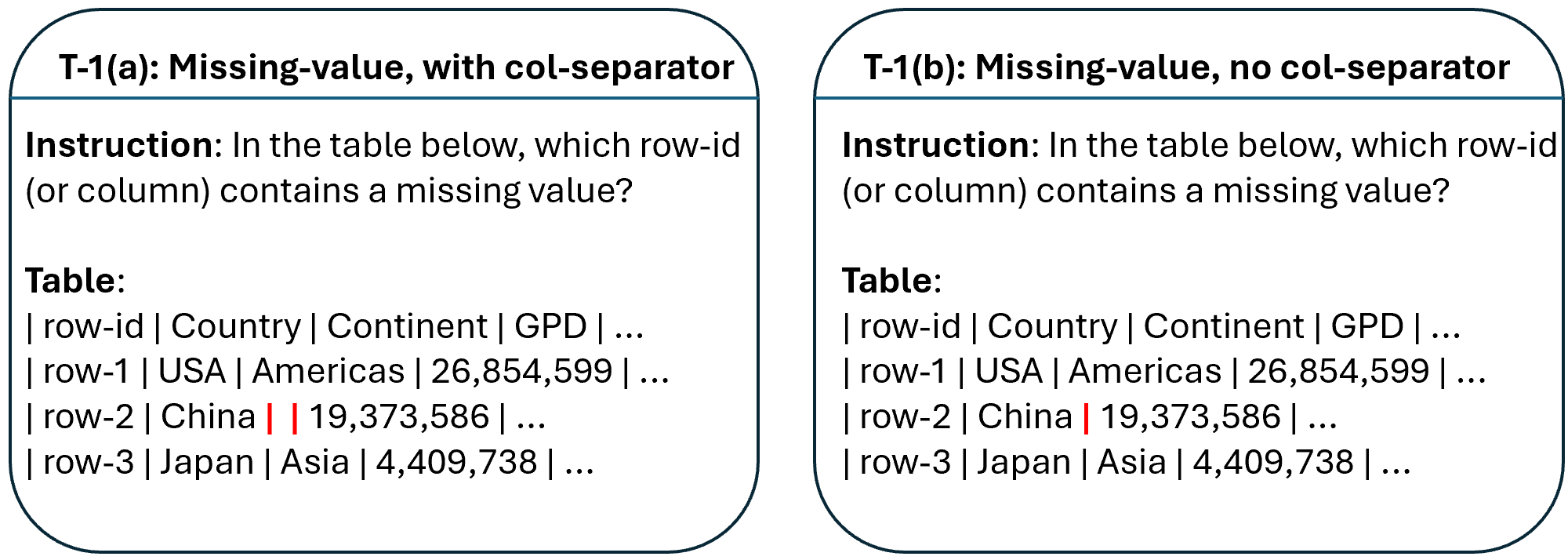}
    \vspace{-6mm}
    \caption{Two variants  of the task (T-1) Missing-cell-identification. (Left) T-1(a): We remove a random cell from a real table, but keep its column-separator. The presence of  {\color{red}``\code{| |}''} indicates a missing cell, which should be easy to identify.  (Right) T-1(b): We remove a random cell, as well as its column-separator, which is a common but challenging CSV parsing issue~\cite{csv-parse-1, csv-parse-2, csv-parse-3}.}
    %\vspace{-5mm}
    \label{fig:task-missing-cell}
\end{figure}

To test language models' ability to read tables in the columnar direction, we design simple tests. In the first test, referred to as ``Missing-value-identification'' (T-1 in Table~\ref{tab:task-summary}), we sample a real table $T$ with no missing cells, and remove a random cell from $T$. We then produce two variants of the test, like shown in Figure~\ref{fig:task-missing-cell}: 

\begin{itemize}[noitemsep,topsep=0pt,leftmargin=*]
\item[] \underline{T-1(a)}:  we keep the column separator of the missing cell and ask language-models to identify the row-id/column-header of the missing cell, like in  Figure~\ref{fig:task-missing-cell} (Left), which seems simple;
\item[] \underline{T-1(b)}: We remove the column separator of the missing cell and then ask the same question, like in Figure~\ref{fig:task-missing-cell} (Right). This is a common situation in CSV parsing that can be challenging~\cite{csv-parse-1, csv-parse-2, csv-parse-3}, as one needs to align values vertically to see the missing value is in which column. (In the case, humans can see that the countries ``\code{USA}'' and ``\code{China}'' should align, the GPD numbers should align, so there must be a missing cell in ``\code{row-2}'', in between ``\code{China}'' and ``\code{19,373,586}'', for the column ``\code{Continent}''). 
\end{itemize}

We repeat these two tests 1000 times, using 1000 randomly sampled real tables.
Table~\ref{tab:result-missing-cell} shows the result of this test. We can see that it is clearly challenging for language models to read tables in the column direction, where the accuracy with and without column-separator is 0.38 and 0.26, respectively. Even with column-separator and explicit few-shot demonstrations, the model is only able to get half of the tests right (0.51). %Similar results are also observed in alternative models, such as Chat-GPT. 

In the row-direction, the model's ability to identify a missing cell is clearly better, though still not great, especially in the ``no col-separator'' setting.

To ensure that the language models are not confused by what we mean in ``missing cell'', we create a second, even simpler test, called Column-Finding (T-2 in Table~\ref{tab:task-summary}), shown with an example in Figure~\ref{fig:basic-tests} (Right), where we ask the model to find the column-header of a specific value, which appears exactly once in a given table $T$, for 1000 randomly sampled real tables. 
Our result show that the accuracy of GPT-3 is similarly low (0.46), confirming the hypothesis that language models ability to read two dimensional tables is likely insufficient. %(Both of our tests above can be replicated simply in OpenAI playgrounds).

\begin{table}
\scriptsize
    \centering
    \begin{tabular}{|c|>{\centering\arraybackslash}p{1.2cm}|>{\centering\arraybackslash}p{1.3cm}|>{\centering\arraybackslash}p{1.2cm}|>{\centering\arraybackslash}p{1.3cm}|}  \hline
           \multirow{2}{*}{ (T-1): Missing cell} &  \multicolumn{2}{c|}{Find col-header tests:}  & \multicolumn{2}{c|}{Find row-id tests:}  \\ \cline{2-5}
         & (a) no col-sep   & (b) has col-sep & (a) no col-sep & (b) has col-sep \\ \hline
        GPT-3.5 (zero-shot) & 0.26 & 0.30 & 0.76 & 0.87 \\ \hline
        GPT-3.5 (few-shot) & 0.38 & 0.51 & 0.77 & 0.91 \\ \hline
        % Chat-GPT (zero-shot) & 0.30 & 0.42 & 0.82 & 0.92 \\ \hline
        % Chat-GPT (few-shot) & 0.46 & 0.63   & 0.85 & 0.96\\ \hline
    \end{tabular}
    \caption{Accuracy numbers of vanilla  GPT-3.5 (using \code{Text-Davinci-002}), on the task of (T-1) Missing-value-identification as shown in~\ref{fig:task-missing-cell}.}
    \vspace{-6mm}
    \label{tab:result-missing-cell}
\end{table}

\underline{Order-sensitive (text) vs. permutation-invariant (tables)}. In addition, we observe that natural-language texts tend to be 
\emph{order-sensitive}, where swapping two tokens will generally lead to different meanings (e.g., ``Jennifer called you'' vs. ``you called Jennifer'').  In comparison, tables tend to be \emph{permutation-invariant}, where swapping two rows or two columns, should generally not change the semantic meaning of the resulting table. 

As a result, when applying language-models to table-tasks, e.g., Entity-matching, Error-Detection, Schema-Matching, we find that the predictions can be rather sensitive to the order in which columns are presented in the input tables, even when we only slightly re-order the columns. 

We believe it shows that language models understanding of tables is still unstable and sub-optimal (likely influenced by the overwhelming text data used in its pre-training),  because the decisions for tasks like Entity-matching and Error-Detection should really not depend on the order of columns. % -- in fact, when we present re-ordered tables to humans for tasks like Entity-matching and Error-Detection, human predictions are almost always stable, showing that language models ability to understand tables are still sub-optimal, likely affected by the overwhelming text data used in its pre-training.

\underline{Other differences.} There are a number of additional aspects that make tables different from text. For example, table-cells tend to be short-form entity-names or phrases, which when serialized in a row, will typically be different from natural-language sentences found in text documents. Furthermore, values in the same column   generally have homogeneous values, with pairs of columns encode regular relationships, which is another property not found in texts. 
All of these make tables different from texts, %yet language-models are pre-trained predominantly on texts, 
likely rendering language-models sub-optimal for table use cases, which motivates our table-tuning approach described next.

%We tried ways to ``hack'' it, e.g., using transposed versions of the input table, with mixed results. Report in Section~\ref{sec:lessons}

%% file: FineTune.tex
\section{Table-tuning for \sys}
\label{sec:table-tune}

\input{TableForAllTasks}

We propose a new table-tuning paradigm, to enhance language models ability to understand tables and perform table-tasks,  %, where we use large amounts of demonstrations of diverse table-tasks, to  so that they can (1) become more optimized to table use cases, while (2) still being general and able to respond to new and unseen tasks. 

\subsection{Overall approach: Synthesis-then-Augment} 
Like discussed earlier, our table-tuning is inspired by the success of ``\emph{instruction-tuning}'' from the NLP literature~\cite{flan, self-instruct, instruct-gpt}, illustrated in Figure~\ref{fig:instruction-tuning-vs-table-tuning} (Left), where diverse training data in the form of ``\code{(instruction, completion)}'' pairs are used to continue to train language-models, and has led to popular models like ChatGPT and LLaMa-chat that can understand and follow human instructions.

%language models like GPT-3  continue to be trained using large amounts of (instruction, completion) pairs, to enhance models' ability to directly follow human instructions (e.g., ``\code{write an essay for college admission}''), without using few-shot examples.  
%This approach is proven to be highly effective, and has led to successful systems such as Instruct-GPT~\cite{instruct-gpt} (the predecessor of Chat-GPT), Chat-GPT, and LLaMa-Chat~\cite{llm-llama}. 

%The success of ``{instruction-tuning}'' is clearly encouraging. 

Our proposed \emph{table-tuning}, as illustrated in Figure~\ref{fig:instruction-tuning-vs-table-tuning} (Right), is similar in spirit -- instead of improving language-model ability to follow instructions using diverse ``\code{(instruction, completion)}'' pairs, we aim to improve language-model ability to perform table tasks using diverse ``\code{(instruction, table, completion)}'' triples, where each such triple defines an instance of a \emph{table-task}:

\begin{definition}
\label{def:table-task}
An instance of a \emph{table-task}, denoted by $t$, is defined as a triple $t =(Ins, T, C)$,  where  $Ins$ is the natural-language instruction that specifies the table-task,  $T$ is the input table on which the task is to be performed, and $C$ is the expected completion from following the instruction $Ins$ and performing the task on table $T$. 
\end{definition}

\begin{example}
\label{ex:table-task}
The examples in Figure~\ref{fig:basic-tests}, Figure~\ref{fig:more-tests}, and Figure~\ref{fig:instruction-tuning-vs-table-tuning}, show simple examples of table-tasks,  defined by the $(Ins, T, C)$ triples, which correspond to (\code{instruction, table, completion}), respectively.  Note that the completion $C$ can be natural-language texts (with JSON or other alternatives for answer parsing), tables, or a combination of both. 
\end{example}

The challenge, however, is that prior work on instruction-tuning have shown that the quality of the ``\code{(instruction, completion)}'' pairs is crucial~\cite{llm-llama, instruct-gpt}, to the extent that companies hired armies of human labelers to manually label such data, (e.g.,  instruction: \code{``write a bed-time story with a bear goes to beach}'', completion: \code{an-actual-story-with-bears})~\cite{instruct-gpt}, to ensure the quality and diverse of the training data. 

We would like to replicate the success of instruction-tuning  in the table domain, but ideally without the expensive human labeling. 

\underline{Reusing existing benchmark data: insufficient diversity.} One approach to generate table-tasks, is to use existing benchmark data published in the database literature (similar efforts were made in the NLP literature for instruction-tuning~\cite{flan}).

However, we found that the existing benchmark data to have:
\begin{itemize}[noitemsep,topsep=0pt,leftmargin=*]
\item[] (1)  \emph{limited task-diversity}: as the literature tends to focus on a few select  table-tasks that are hard and challenging (e.g., entity-matching and data-transformation); and 
\item[] (2)  \emph{limited data-diversity}: as benchmark data are typically labeled manually by researchers, only on a few specific datasets, which is sufficient for benchmark evaluation purposes, but insufficient when we want to use them as ``training data'' for language models.  
\end{itemize}

\noindent Our attempt to use only existing benchmark data for table-tuning leads to over-fitting, due to the lack of task and data diversity. %(Section~\ref{sec:lessons}). 

\underline{Our approach: Synthesis-then-Augment.} We therefore propose a ``\emph{synthesize-then-augment}'' approach to create diverse table-tasks using real tables, which can be used as training-data to table-tune language-models.

We show the main steps of our synthesize-then-augment approach in Algorithm~\ref{alg:main}. First, we sample a table $T \in \mathbf{C}$ from a large corpus of real tables $\mathbf{C}$,  and a type of table-task $S \in \mathbf{S}$. From the $(T, S)$ pair, we synthesize an instance of a table-task $t =(Ins, T, C)$ (line~\ref{line:synthesis}), which is the task-synthesis step we will discuss in detail in Section~\ref{sec:synthesize-tasks}.   From the set of diverse instances of table-tasks created $(Ins, T, C)$, we then proceed to ``augment'' the tasks, at instruction/table/completion levels (line~\ref{line:augment-ins}-\ref{line:augment-completion}), which is a step that we will describe in Section~\ref{sec:augment-tasks}. The resulting table-tasks $A = \{(Ins', T', C')\}$ become the training data we use to table-tune language-models.

\begin{small}
\begin{algorithm}[t]
\SetKw{kwReturn}{return}
 \Input{A corpus of diverse real tables $\mathbf{C}$, a set of table-task types $\mathbf{S}$}
 \Output{Diverse synthesized table-tasks $A = \{(Ins, T, C)\}$}
 
 %$V \leftarrow \{v_T | T \in \mathbf{T} \}$, with $v_T$ representing each $T \in \mathbf{T}$ %\tcp{Algorithm~\ref{algo:constraints}} 

 $D \leftarrow \{\}, A \leftarrow \{\}$
 
  \ForEach{$T \in \mathbf{C}, S \in \mathbf{S}$   \label{line:iterate}}  
    {
       $(Ins, T, C) \leftarrow \text{Synthesize-Table-Task}(S, T)$ {\footnotesize \tcp{(Section~\ref{sec:synthesize-tasks})}} \label{line:synthesis} 
       
      $D \leftarrow D \cup (Ins, T, C)$

    }

\ForEach{$(Ins, T, C) \in D$ \label{line:1mca-IND}}  
{
       $Ins‘ \leftarrow \text{Augment-Instruction}(Ins)$ {\footnotesize \tcp{(Section~\ref{sec:augment-tasks})}}  \label{line:augment-ins} 
       
       $T‘ \leftarrow \text{Augment-Table}(T)$ {\footnotesize \tcp{(Section~\ref{sec:augment-tasks})}}  \label{line:augment-table} 
       
       $C‘ \leftarrow \text{Augment-Completion}(C)$ {\footnotesize \tcp{(Section~\ref{sec:augment-tasks})}}  \label{line:augment-completion}

       $A \leftarrow A \cup (Ins', T', C')$
}

\kwReturn $A$
\caption{Synthesize table-tasks for table-tuning}
\label{alg:main}
\end{algorithm}
\end{small}

\subsection{Synthesize diverse table-tasks} 
\label{sec:synthesize-tasks}

We now describe how we synthesize diverse instances of table-tasks $t=(Ins, T, C)$ (Line~\ref{line:synthesis} of Algorithm~\ref{alg:main}), so as to exercise language-models ability to understand two-dimensional table structures. 

We propose two complementary approaches that (1) synthesize new table-tasks for task-diversity, and (2) synthesize new table test-cases of existing table-tasks for data-diversity. We will discuss each below in turn.
%we believe that it is not necessary to focus only on the challenging tasks popular in the literature.  synthesize tasks, using diverse and real tables. 

%instruction / input-table / output (an entire table, or an answer in json)

\textbf{Synthesize new table-tasks for task-diversity.} Since our goal is to enhance language models' ability to understand tables, we believe it is not necessary to focus exclusively on challenging table-tasks that have been the focus of the literature~\cite{data-cleaning-survey}. Instead, we propose a number of table-understanding/augmentation/manipulation tasks that are easy to synthesize, leveraging large amounts of real tables that already exist.  Specifically, we crawled 2.9M high-quality web-tables (e.g., Wikipedia)~\cite{bing-assets}, referred to as $\mathbf{C}^{wt}$, and 188K database-tables (extracted from BI data models)~\cite{auto-bi}, referred to as $\mathbf{C}^{db}$, and synthesize table-tasks based on real tables sampled from the corpus.

We will go over the list of synthesized table-tasks below:

\underline{(T-13) Table summarization (TS).} Since web-tables often have descriptive titles, we synthesize a table-summarization task, where we ask the model to summarize the content in a table. Specifically, we sample $T \in \mathbf{C}^{wt}$ whose extracted table-title $title(T)$ are neither too long nor too short, and create a table-summarization task as: 
\begin{equation*}
TS(T) = (Ins^{TS}, T, title(T))
\end{equation*}
where $Ins^{TS}$ is the canonical human-instruction to describe the TS task (e.g., ``\code{Please provide a succinct summary for the table below}''), which we will further augment for diversity (Section~\ref{sec:augment-tasks}), $T$ is the input table we sampled from $\mathbf{C}^{wt}$, and $title(T)$  is its expected completion.

This task is designed to use real tables, with real human annotated titles, to enhance models ability to read tables and understand the highlights from the table. Note that although we use $title(T)$  as the expected completion/answer, it does not over-constrain language-models to over-fit on such answers -- it only nudges language-models in that general direction, just like training data in the form of (\code{``write a bed-time story with a bear goes to beach}'' $\rightarrow$ \code{an-actual-human-written-story}) used in instruction-tuning does not over-constrain/over-fit the underlying models.

\underline{(T-14) Column augmentation.} 
Since we have lots of real tables in $\mathbf{C}^{wt}$ and $\mathbf{C}^{db}$, we create a task where we take the first $k$ columns in a table $T$, denoted as  $C_{[1, k]}(T)$, and ask the language-models  to generate the $(k+1)$-th column $C_{k+1}(T)$, written as follows:
\begin{equation*}
CA(T, k) = (Ins^{CA}, C_{[1, k]}(T), C_{k+1}(T))
\end{equation*}
where $Ins^{CA}$ is again the natural-language instruction that describes the row-augmentation task. 
This task exercises a model's ability to generate realistic columns given a table context that need to be semantically compatible.

\underline{(T-15) Row augmentation (RA).} 
Similar to Column-augmentation, we synthesize a Row-augmentation task where we sample a table $T$ and ask the model to generate the $(k+1)$-th row, given the first $k$ rows, written as:
\begin{equation*}
RA(T, k) = (Ins^{RA}, R_{[1, k]}(T), R_{k+1}(T))
\end{equation*}
This task exercises a model's ability to synthesize realistic rows given a table context, which need to  align vertically with existing rows.

\underline{(T-16) Row/column swapping (RS/CS).} 
In this task, we ask the models to perform a table-manipulation step, where given a sampled table $T$, we provide an instruction to swap the $i$-th and $j$-th row. We programmatically generate the resulting output table from the swap operation,  denoted as $Swap(T, R_i, R_j)$, which is the target ``\code{completion}''. The Row-swapping task $RS_{i, j}(T)$ is written as:
\begin{equation*}
RS_{i, j}(T) = (Ins^{RS}, T, Swap(T, R_i, R_j))
\end{equation*}
We similarly synthesize the Column-swapping task $CS_{i, j}(T)$  as:
\begin{equation*}
CS_{i,j}(T) = (Ins^{CS}, T, Swap(T, C_i, C_j))
\end{equation*}
We note that tasks like Row/Column-swapping would seem simple to perform, both programmatically or through UI interactions (e.g., inside spreadsheets using menu options), and are therefore not tasks studied in the  literature (unlike more challenging tasks like entity-matching or data-transformation). We are similarly not intending to use table tasks as ``tests'', but because ``tables serialized as natural-language texts'' are ultimately the only way to feed input into language models (regardless of whether we want to output to be text/code/table/etc.), these table-tasks are still useful as ``training data'' for models to better read and understand tables.

\underline{(T-17) Row/column filtering.} 
In this table-manipulation task, we ask models to filter down to specific rows/columns on a sampled table $T$, based on a specified set of row/column indexes $S$:  
\begin{equation*}
RF_S(T) = (Ins^{RF}, T, R_S(T))
\end{equation*}
\begin{equation*}
CF_S(T) = (Ins^{CF}, T, C_S(T))
\end{equation*}
These tests are again meant to exercise model ability to manipulate tables, where cells in both vertical and horizontal directions need to be aligned. 

\underline{(T-18) Row/column sorting (RS/CS).} 
In the sorting tasks, we ask models to sort rows in a table $T$, based on values in a column $C$, where the expected output table can be programmatically generated, which we write as $Sort_C(T)$, so that the task $RS_C(T)$ is: 
\begin{equation*}
RS_C(T) = (Ins^{RS}, T, Sort_C(T))
\end{equation*}
Similarly, we have a task to sort columns in a table $T$, based on column-headers $H$, written as  $CSs(T)$:
\begin{equation*}
CS(T) = (Ins^{CS}, T, Sort_H(T))
\end{equation*}
We note that the sorting tasks are fairly challenging for language-models -- while we do not expect models to be perfect on such tasks, they exercises model ability to manipulate tables nevertheless. 

\underline{(T-11) Head-value matching (HVM).} 
In this task, we sample a table $T$, remove all its column headers $H$ to produce the corresponding table without headers, $\overline{T}$. We then shuffle these headers $H$, and ask models to fill $H$  into $T'$, to produce the $\text{HVM}(T)$ task: 
\begin{equation*}
\text{HVM}(T) = (Ins^{HVM}, \overline{T}, T)
\end{equation*}
Like other tasks above, HVM is another task that we can synthesize in large quantities, using real tables, and without labeling. It is intended to be a task that helps models to better understand and correlate the semantics of column-headers and values.
 
\underline{Discussions.} We show in our experiments, that using synthesized table-tasks on diverse tables improves the task- and data-diversity, which lead to better model generalizability (our ablation study shows that without these synthesized tasks there is a substantial drop in model quality). 

Our list of synthesized table-tasks, however, is obviously not meant to be exhaustive, and is only a starting point. We believe that with some creativity, many more tasks can be synthesized  to further improve the table-tuning process. For comparison, the NLP community has amassed over 1000 tasks for instruction-tuning, in a community effort~\cite{flanscaling}, where they show that having more and diverse tasks always helps instruction-tuning.

%Note that our basic tests like (T-1) missing-cell-identification and (T-2) column-finding (Table~\ref{tab:task-summary} and with examples in Figure~\ref{fig:basic-tests}) are synthesized and can be used as training data too. However, we intentionally leave out these tasks, as hold-out tasks unseen during table-tuning, so that we can evaluate models ability in understanding tables in our experiment evaluations (Section~\ref{sec:exp}).

\textbf{Synthesize new table test-cases for data-diversity.}
There are a number of existing and important table-tasks, such as data-transformation, entity-matching, etc. that are extensively studied in the database literature. We want to use these established tasks in table-tuning too, also in the ``\code{(instruction, table, completion)}'' format. However, like mentioned earlier, the existing benchmarks for these tasks are typically manually labeled on a few datasets, which can be used to evaluation, but are unfit  as training data for table-tuning, due to their limited quantities and diversity.

Instead, we synthesize new table test-cases for these established table-tasks, using real tables sampled from $\mathbf{C}^{wt}$ and $\mathbf{C}^{db}$.

\underline{(T-5) Row-to-row Data Transformation (R2R)}~\cite{data-transform-flashfill, data-transform-tde}. To synthesize diverse test tables with data-transformations, we run a production-quality 
\iftoggle{fullversion}
{
program-synthesizer~\cite{data-transform-tde}, 
}
{
program-synthesizer, 
}
on web-tables sampled from $\mathbf{C}^{wb}$, to identify tables $T\in \mathbf{C}^{wb}$ where some columns $C_{in} \subset T$ can be transformed into $C_{out}  \subset T$, using an inferred program $P$, such that $P(C_{in}) = C_{out}$ hold on all rows in $T$ (e.g., \code{(first-name, last-name)} $\rightarrow$ \code{(full-name)} in the same table
    \iftoggle{fullversion}
    {
    \cite{auto-transform}).
    }
    {
    ).
    }
We then remove one random value $v \in C_{out}$ from $T$, to produce a test table $T_{-v}$ where $v$ is missing. We then synthesize a task $R2R(T)$:
\begin{equation*}
R2R(T) = (Ins^{R2R}, T_{-v}, T)
\end{equation*}
where given $T_{-v}$ as the input, we want to the model to infer the transformation and fill in the missing $v$ to produce $T$.

\underline{(T-7) Schema Matching (SM)}~\cite{schema-mapping-survey}. To synthesize new table test cases for schema matching, we sample a real table $T$, and take the first $k$ rows of $T$ to produce $T_1 = R_{[1, ~k]}(T)$. We then take the next $k$ rows from $T$ to produce $T_2 = R_{[k+1,~2k]}(T)$, where we additionally ``paraphrase'' the column-headers of the original $T$, into new column-headers in $T_2$, using a mapping of semantically-similar column-names  generated by GPT, denoted as $M$ (e.g., ``\code{company names}'' $\rightarrow$ ``\code{enterprises}'', ``\code{emp-id}'' $\rightarrow$ ``\code{employee identifier}'', etc.). Finally, we shuffle the columns in $T_1$ and $T_2$, and make the two a test case for schema matching, where the ground-truth is in $M$. The resulting task is written as $SM(T)$:
\begin{equation*}
SM(T) = (Ins^{SM}, (T_1, T_2), M)
\end{equation*}
This again can systematically generate large numbers of schema-matching test tables, as training data for table-tuning.

\underline{(T-8) Data Imputation (DI)}~\cite{imputation-1, imputation-2}. For data imputation, we randomly sample a real table $T$, and then remove a random value $v \in T$, to produce $T_{-v}$. The task $DI(T)$ is then to predict the missing $v$ from its table context:
\begin{equation*}
DI(T) = (Ins^{DI}, T_{-v}, v)
\end{equation*}
Note that while not all missing values $v$ in DI tasks so generated can be reliably predicted, it nevertheless exercises models' ability to leverage correlations that exist between values in the row and column contexts.

\underline{(T-9) Error Detection (ED)}~\cite{data-cleaning-survey}. To synthesize error-detection tasks, we sample a real table $T \in \mathbf{C}^{wt}$, and generate a modified $\Tilde{T}$, where we replace a value $v\in T$ with $v'$, using an existing package~\cite{typo-gen} that injects one likely typographic error into $v$. The task $ED(T)$ is then:
\begin{equation*}
ED(T) = (Ins^{ED}, \Tilde{T}, v')
\end{equation*}
where we aim to identify the misspelled $v' \in \Tilde{T}$ based on surrounding table context. 

\underline{(T-10) List extraction (LE)}~\cite{list-extraction-google, list-extraction-tegra}. To synthesize the task of extracting tables from list data without explicit column-delimiters, we sample a table $T$, and replace all column separators with white spaces to generate its unsegmented list-form $L(T)$. The task $LE(T)$ is then:
\begin{equation*}
LE(T) = (Ins^{LE}, L(T), T)
\end{equation*}
which is to produce the correct column-segmentation of $L(T)$, and generate the corresponding table $T$, based on value alignment in the vertical direction.

Since we have large numbers of diverse tables, in Line~\ref{line:synthesis} of Algorithm~\ref{alg:main} we make sure that each table $T$ is used by one task-type above, to synthesize one instance of table-task, to ensure the diversity of data we generate.

%We compile and standardize the relevant datasets for tasks of interest, such as Entity-Matching, Schema-Matching, and Row-to-row Data Transformation, like summarized in Table~\ref{tab:task-summary}. We use these datasets primarily for testing only, because the diversity and quantity of these tasks (e.g., a few hundred test cases per task) are often not enough to produce and test at the same time. 

%\textbf{Leverage existing tasks and benchmark data}

%No models in the database literature, aim to cover such a diverse range of table-tasks. 

\subsection{Augment synthesized table-tasks} 
\label{sec:augment-tasks}
From synthesized instances of table-tasks $t = (Ins, T, C)$, we then perform additional augmentations at multiple levels, corresponding to steps in Line~\ref{line:augment-ins}-Line~\ref{line:augment-completion} of Algorithm~\ref{alg:main}, where the goal is to create even more task/data diversity and avoid over-fitting in table-tuning. 

We will go over different levels of augmentations below in turn.

\textbf{Instruction-level augmentations.} At the instruction level, because using the same instruction repeatedly across training-data instances can lead to over-fitting~\cite{self-instruct}, we augment the canonical instruction using generative models like GPT to paraphrase the canonical human-written instruction into many different variants. 

For example, for the task-type (T-13): Table-Summarization (Section~\ref{sec:synthesize-tasks}),  the canonical human-written instruction is:  ``\code{Please look at the table below and provide a title that can summarize the table}''. We generate alternative instructions for the task using language-models, to produce variations such as ``\code{Please examine the table below and give it a descriptive title}'', in a manner similar to~\cite{self-instruct}, which we can then use to populate instances of table-tasks as alternative instructions (Line~\ref{line:augment-ins}).

\textbf{Table-level augmentations.} At the table-level, we know that two-dimensional tables should largely be ``permutation-invariant'', where permutating rows and columns should generally lead to a table with similar semantic meanings
(Section~\ref{sec:read_tables}), at the table-level we can perform operations such as  column-permutation, row-permutation, column-sampling, and row-sampling, to increase the diversity of tables used in our table tasks. 

When the training data has an original instance of the table-task, $t = (Ins, T, C)$, and its augmented version $t' = (Ins, T', C)$, where $T'$ is an augmented version of $T$, which has the same semantic meaning and thus the same completion $C$, the hope is that by continuing to train language-models on such training-data, we can increase model stability on tables and make them less sensitive to ``semantic-preserving table-operations'' (e.g., column-reordering like discussed in Section~\ref{sec:read_tables}).

\begin{figure}[t]
    \hspace{-2mm}
    %\centering  
    \includegraphics[width=1.05\columnwidth]{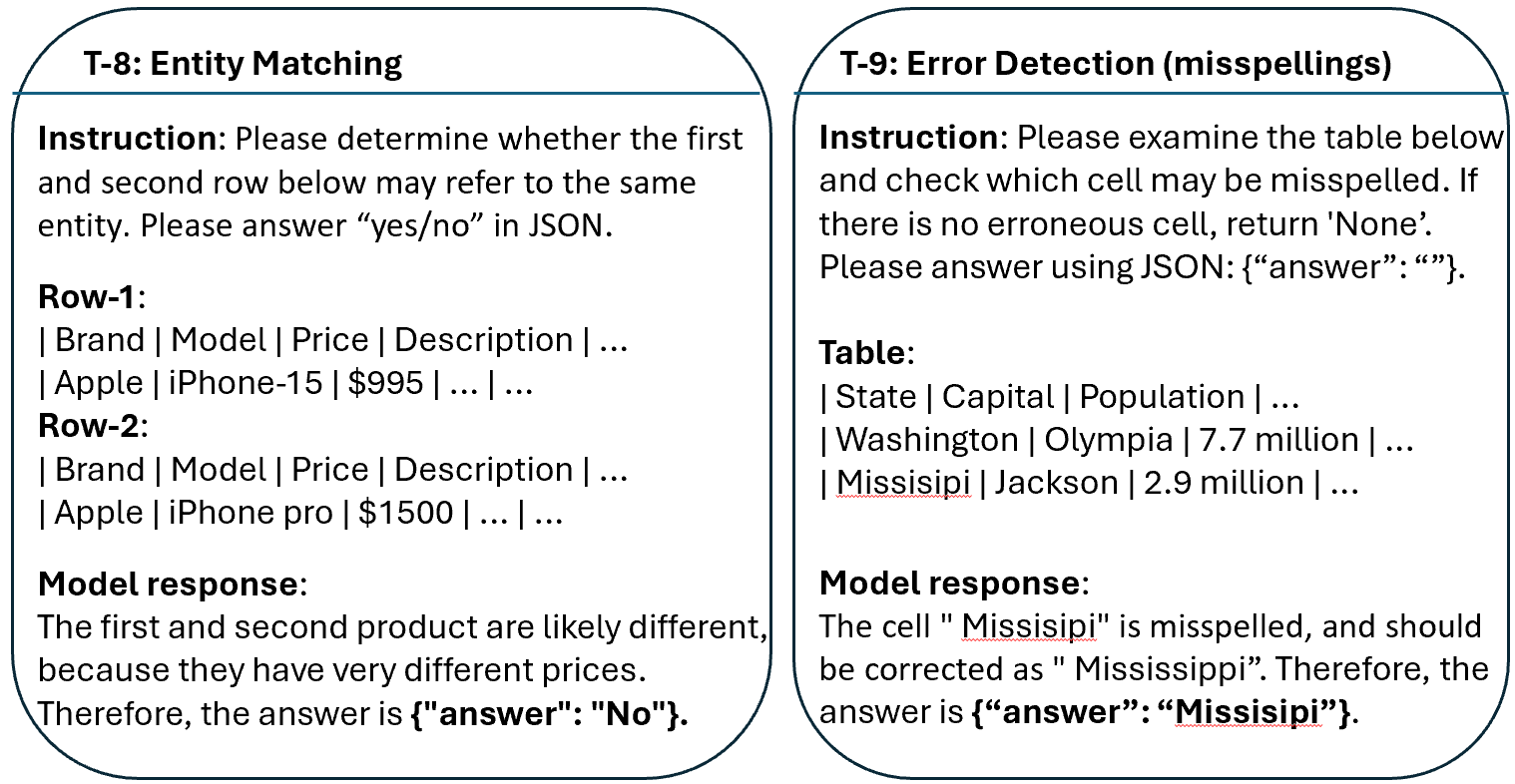}
    \vspace{-2mm}
    \caption{Example table-tasks we generate for (T-8) Entity-matching, and (T-9) Error-detection, using ``augmented-completions'' that contain reasoning steps similar to chain-of-thought, which when used as training-data in table-tuning, can ground model responses and improve result quality.}
    %\vspace{-5mm}
    \label{fig:task-cot}
\end{figure}

\textbf{Completion-level augmentations.} At the completion-level, for a synthesized instance of table-task $t=(Ins, T, C)$, we augment our original completion $C$ by generating $C'$ that adds reasoning steps into the original $C$,
after we observe that performing reasoning-steps (similar to~\cite{cot}) on more complex table-tasks (e.g., entity-matching and error-detection) can lead to better task performance. 

\underline{Language-model assisted completion-augmentation.}
We show the completion-level augmentation using examples in Figure~\ref{fig:task-cot}. The left of the figure is the task of (T-8) Entity-matching, where  models need to identify whether two table-rows may refer to the same real-world entities. We tried two modes of operations on vanilla language-models (before table-tuning):
\begin{itemize}[noitemsep,topsep=0pt,leftmargin=*]
\item[] (1)  \emph{Direct-answer}: where models are asked to produce a ``yes/no'' answer directly, without additional reasoning;
\item[] (2)  \emph{Reason-then-answer}: where we require the models to ``explain step-by-step''  in the prompt, before producing a ``yes/no'' answer.
\end{itemize}
We observe that for complex tasks like Entity-matching, models in the ``direct-answer'' mode are more prone to make mistakes (e.g., answering with only a ``no'' in the response), leading to lower accuracy; whereas with ``reason-then-answer'', even vanilla language-models perform better,  likely because the step-by-step reasoning chain makes it likely to generate the correct answer, similar to what was observed in the NLP literature (e.g., chain-of-thought reasoning~\cite{cot} and self-consistency~\cite{self-consistency}, etc.).

Recall that our goal is to table-tune language models, so that the resulting model can perform better on complex table-tasks, by hopefully developing a habit of performing step-by-step reasoning. In order to produce training-data for table-tuning, so that language-models can learn-to-reason on complex tasks, we augment $t=(Ins, T, C)$ by replacing the direct yes/no answer $C$, with a more detailed answer using step-by-step reasoning (Figure~\ref{fig:task-cot} shows a simplified example), which we generate by providing GPT with the yes/no answer, and then ask the model to produce reasoning-steps as our new $C'$. We find this to encourage  table-tuned models to perform correct reasoning-steps, on other complex table-tasks that are similar in nature.

\underline{Ground-truth assisted completion-augmentation.}
The right of Figure~\ref{fig:task-cot} shows another example of completion-augmentation, where we use ground-truth to produce step-by-step reasoning, for the example task of (T-9): Error-detection. The task here is to identify misspelled cells from a given table. We find that vanilla language-models  to produce lots of false-positives on the error-detection task, where the models confidently predict a cell-value (e.g., an abbreviation or uncommon company/person/location names) to be misspelled, when no such misspellings exist upon close inspection. The tendency to produce false-positives persists, with or without ``step-by-step'' chain-of-thought style prompting.  

Because we cannot rely on chain-of-thought style prompts to generate correct reasoning steps as augmented-completions $C'$, we instead use the ground-truth available to us in the table-task synthesis step (Section~\ref{sec:synthesize-tasks}), to generate augmented completions that embeds the reasoning step. Specifically, like shown in Figure~\ref{fig:task-cot} (right), we augment the completion to contain not only the prediction (a value $v$ is misspelled), but also generate the explanation that points out the correct version of the predicted misspelling (e.g., ``\code{Missisipi}'' should be ``\code{Mississippi}''). We find this grounds the language-models predictions with an actual explanation, which substantially reduces false-positives and improve result quality.

\textbf{Additional augmentations.} Along the lines of augmentations, there are additional types of augmentations we perform, including ``\emph{template-level augmentation}'', where we mix zero-shot task template and few-shot task template (which appends multiple input-table/output-completion examples after the instruction $Ins$), as well as ``\emph{task-level augmentation}'' (by synthesizing new types of table-tasks), which all improve training-data diversity and help table-tuning.
%\textbf{Template-level.} mix zero-shot template and few-shot template.

%\textbf{Task-level.} synthesized tasks

\subsection{\sys as ``table foundation models''} 
\label{sec:table-foundation-model}
Using the synthesis-then-augment approach in Algorithm~\ref{alg:main}, describe in previous sections, we now generate large numbers of diverse table-tasks $A = \{(Ins, T, C)\}$.  We then continue to train language models such as GPT, using serialized $(Ins, T)$ as the ``prompt'', and $C$ as the ``completion'', where we minimize the language-modeling loss of completion given the prompt, subject to regularization. We refer to this  process as table-tuning.

Let $M$ be a decoder-style language model, such as GPT and ChatGPT, let $\text{TableTune}(M)$ be the table-tuned version of $M$. We argue that $\text{TableTune}(M)$ could serve as a better ``table foundation model'', if it performs better than $M$ on table-tasks, in all of the following scenarios:
\begin{itemize}[noitemsep,topsep=0pt,leftmargin=*]
\item[] (1)  \underline{Out of the box zero-shot}: when we use only instructions for $M$ or $\text{TableTune}(M)$ to perform table-tasks;
\item[] (2)  \underline{Out of the box few-shot}: when we use instructions and \emph{randomly selected} few-shot examples to perform table-tasks;
\item[] (3)  \underline{Task-specific prompt-tuning}: when we have a small amount of labeled data for a downstream task, and perform prompt-tuning to select the best instruction/example combinations;
\item[] (4)  \underline{Task-specific fine-tuning}: when we have sufficient amounts of labeled data, and perform task-specific fine-tuning for a task.
\end{itemize}

If table-tuning is effective for language models to learn to better understand and manipulate tables, we expect that $\text{TableTune}(M)$ can perform better on most if not all of the scenarios described above, which is the goal of our experimental evaluation next.

%% file: TableForAllTasks.tex
\begin{table*}
\small
   % \centering
%\hspace{-3mm}
    \begin{tabular}{|p{1.7in}|>{\centering\arraybackslash}p{2.2in}|>{\centering\arraybackslash}p{1.1in}|>{\centering\arraybackslash}p{0.6in}|>{\centering\arraybackslash}p{0.5in}|} \hline
        \multicolumn{1}{|c|}{\textbf{Task-name}} & \textbf{Task description (related work)}  & \textbf{Task category} & \textbf{Table data}  & \textbf{Train/Test} \\ \hline
         T-1: Missing-value identification (MV) & Identify the row and column position of the only missing cell in a given table  & Table understanding &  synthesized  & Test only \\ \hline 
         T-2: Column-finding (CF) & Identify the column-name of a specific value that appears only once in a given table & Table Understanding &  synthesized  & Test only \\ \hline
         T-3: Table-QA (TQA) & Answer a natural-language question based on the content of a table (\cite{table-qa-2, table-qa-tabfact, table-qa-wikitablequestions})  & Table QA &  \cite{table-qa-wikitablequestions}  & Test only \\ \hline
         T-4: Column type annotation (CTA) & Find the semantic type of a column, from a given list of choices (\cite{cta-autotype, cta-sherlock, turl})  & Table understanding &  \cite{cta-sherlock, turl}  & Test only \\ \hline  \hline
         T-5: Row-to-row transform (R2R) & Transform table data based on input/output examples (\cite{data-transform-tde, data-transform-flashfill, data-transformation-wrangler}) & Data transformation  & synthesized (test: \cite{data-transform-tde}) & Train/Test \\ \hline 
         T-6: Entity matching (EM) & Match rows from two tables that refer to the same real-world entity (\cite{em-book, em-deepmatcher, em-ditto, auto-em})  & Table matching  & \cite{em-data-magellan} & Train/Test \\ \hline
         T-7: Schema matching (SM) & Match columns from two tables that refer to the same meaning (\cite{schema-mapping-survey, schema-matching-cupid, schema-matching-valentine}) &  Table matching & synthesized (test: \cite{schema-matching-valentine})  & Train/Test \\ \hline
         T-8: Data imputation (DI) & Predict the missing values in a cell based on the table context (\cite{imputation-1, imputation-2}) & Data cleaning & synthesized & Train/Test \\ \hline
         T-9: Error detection (ED) & Detect data values in a table that is a likely error from misspelling (\cite{data-cleaning-survey, error-detection-survey}) &  Data cleaning & synthesized & Train/Test \\ \hline  \hline
         T-10: List extraction (LE) & Extract a structured table, from a list that lacks explicit column delimiters \cite{list-extraction-2, list-extraction-tegra, list-extraction-google} & Data transformation & synthesized & Train only \\ \hline
         T-11: Head value matching (HVM) & Match column-headers with its data values drawn from the same table & Table matching & synthesized & Train only \\ \hline
         T-12: Natural-language to SQL (NS) & Translate a natural-language question on a table into a SQL query (\cite{nl2sql-spider, nl2sql-2}) & NL-to-SQL & \cite{nl2sql-spider} & Train only \\ \hline
         T-13: Table summarization (TS) & Produce a natural-language summary for the content in a table 
         %(\cite{table-summary-2, table-summary-google, table-summary-3}) 
         & Data augmentation & synthesized & Train only \\ \hline
         T-14: Column augmentation (CA) & Augment a table with additional columns compatible with a given table %(\cite{table-augmentation-2, table-augmentation-infogather}) 
         & Data augmentation & synthesized & Train only \\ \hline         
         T-15: Row augmentation (RA) & Augment a table with additional rows compatible with a given table %(\cite{table-augmentation-2, table-augmentation-infogather}) 
         & Data augmentation & synthesized & Train only \\ \hline
         T-16: Row/column swapping (RCSW) & Manipulate a given table, by swapping the position of two rows or columns &  Table manipulation & synthesized  & Train only \\ \hline
         T-17: Row/column filtering (RCF) & Manipulate a given table, by filtering on given rows or columns &  Table manipulation & synthesized  & Train only \\ \hline
         T-18: Row/column sorting (RCS) & Manipulate a given table, by performing sorting on given rows or columns & Table manipulation & synthesized & Train only \\ \hline
    \end{tabular}
    \caption{A summary of 18  table-related tasks, which we collect and synthesize, in order to ``table-tune'' GPT into \sys. [Task categories]: These tasks cover diverse areas such as: table understanding, table-QA, table matching, table cleaning, table transformation, etc. Some of these tasks (T-1 to T-4) are used as unseen hold-out tasks, to evaluate \sys ability to generalize to completely new and unseen tasks. [Table Data]: we choose to ``synthesize'' table tasks from diverse real tables when possible (e.g., when ground-truth can be produced automatically), to ensure the diversity of the training data and avoids over-fitting. When the ground-truth cannot be automatically produced (e.g., entity-matching, table-QA, NL-to-SQL, etc.), we use existing benchmark data from the literature.
    }
    \label{tab:task-summary}
\end{table*}

%% file: experiment.tex
\section{Experiments}
\label{sec:exp}
We perform extensive experiments to evaluate table-tuned GPT relative to vanilla GPT on diverse table tasks. We plan to release our code and data after internal reviews\textsuperscript{1}  \footnotetext{\textsuperscript{1}: \url{https://aka.ms/table-gpt}}.

\begin{table*}[t]
\scalebox{0.8}{
\begin{tabular}{|c|c|c||c|c||c|c||c|c||c|c||}
\hline
% first header line
\multirow{2}{*}{\textbf{Task Type}} & \multirow{2}{*}{\textbf{Task}} & \multirow{2}{*}{\textbf{Dataset}} & 
 \multicolumn{2}{c||}{\textbf{Zero-Shot}}  & \multicolumn{2}{c||}{\textbf{Few-Shot}}  & 
 \multicolumn{2}{c||}{\textbf{Zero-Shot}}  & \multicolumn{2}{c||}{\textbf{Few-Shot}}  \\ \cline{4-11} 
% second header line
& & & GPT-3.5  & +table-tune & GPT-3.5 & +table-tune & ChatGPT  & +table-tune & ChatGPT & +table-tune \\ \hline
% content 
\multirow{10}{*}{Unseen} &  Column Finding & Spreadsheets-CF & 0.461 & \textbf{0.713}  & 0.682 & \textbf{0.816} & 0.699 & \textbf{0.807} & 0.803  & \textbf{0.848} \\ \cline{2-11} 
 & \multirow{4}{*}{Column Type Annotation} & Efthymiou & 0.757 & \textbf{0.886} & 0.784 & \textbf{0.847} & 0.823 & \textbf{0.882}  & 0.806 &  \textbf{0.861} \\ \cline{3-11} 
 &  & Limaye & 0.683 & \textbf{0.755} & 0.719 & \textbf{0.853} & 0.742 & \textbf{0.769} & 0.832 & \textbf{0.853} \\ \cline{3-11} 
 &  & Sherlock & 0.332 & \textbf{0.449} & 0.528 & \textbf{0.538} & 0.454 & \textbf{0.482} & 0.521  & \textbf{0.553} \\ \cline{3-11} 
 &  & T2D & 0.776 & \textbf{0.875} & 0.83 & \textbf{0.915} & 0.827 & \textbf{0.886} & 0.853 & \textbf{0.912} \\ \cline{2-11} 
 & \multirow{4}{*}{Missing Value Identification} & Column (no separator) & 0.261 & \textbf{0.294} & 0.383 & \textbf{0.441} & 0.299 & \textbf{0.351}  & 0.468 & \textbf{0.474} \\ \cline{3-11} 
 &  & Column (with separator) & 0.305 & \textbf{0.457} & 0.519 & \textbf{0.643} & 0.422 & \textbf{0.520}  & 0.635 & \textbf{0.665} \\ \cline{3-11} 
 &  & Row (no separator) & 0.768 & \textbf{0.851} & 0.774 & \textbf{0.882} & 0.822 & \textbf{0.840}  & 0.859 & \textbf{0.894} \\ \cline{3-11} 
 &  & Row (with separator) & 0.875 & \textbf{0.959} & 0.917 & \textbf{0.976} & 0.923 & \textbf{0.936} & 0.960 & \textbf{0.968} \\ \cline{2-11} 
 & Table Question & Wiki & 0.45 & \textbf{0.486} & 0.454 & \textbf{0.478} & 0.512 & \textbf{0.521} & 0.520 & \textbf{0.527} \\ \hline
 
\multirow{17}{*}{Seen} & Data Imputation & Spreadsheets-DI & 0.423 & \textbf{0.558} & 0.57 & \textbf{0.625}  & 0.524 & \textbf{0.594} & 0.609 & \textbf{0.649} \\ \cline{2-11} 
 & \multirow{7}{*}{Entity Matching} & Amazon-Google & 0.153 & \textbf{0.657} & 0.659 & \textbf{0.676} & 0.239 & \textbf{0.566} & 0.680 & \textbf{0.701}  \\ \cline{3-11} 
 &  & Beer & 0.5 & \textbf{0.727} & 0.815 & \textbf{0.923} & 0.741 & \textbf{0.923} & 0.783 & \textbf{0.963}  \\ \cline{3-11} 
 &  & DBLP-ACM & 0.402 & \textbf{0.847} & \textbf{0.954} & 0.912 & 0.833 & \textbf{0.932} & \textbf{0.961} & 0.938 \\ \cline{3-11} 
 &  & DBLP-GoogleScholar & 0.206 & \textbf{0.861} & 0.809 & \textbf{0.896} & 0.632 & \textbf{0.912} & 0.823 & \textbf{0.924} \\ \cline{3-11} 
 &  & Fodors-Zagats & 0.083 & \textbf{0.872} & 0.872 & \textbf{0.977} & 0.809 & \textbf{1.000} & 0.872 & \textbf{0.977} \\ \cline{3-11} 
 &  & Walmart-Amazon & 0.268 & \textbf{0.691} & 0.519 & \textbf{0.711} & 0.206 & \textbf{0.678} & 0.664 & \textbf{0.824} \\ \cline{3-11} 
 &  & iTunes-Amazon & 0 & \textbf{0.788} & 0.826 & \textbf{0.943} & 0.393 & \textbf{0.862} & 0.833 & \textbf{0.929} \\ \cline{2-11} 
 & \multirow{2}{*}{Error Detection} & Spreadsheets-Real & 0.058 & \textbf{0.565} & 0.319 & \textbf{0.552} & 0.058 & \textbf{0.544} & 0.443 & \textbf{0.551} \\ \cline{3-11} 
% &  & ExcelSynthetic & 0.036 & \textbf{0.676} & 0.415 & \textbf{0.813} \\ \cline{3-7} 
 &  & WebTables-Real & 0.077 & \textbf{0.643} & 0.338 & \textbf{0.545} & 0.078 & \textbf{0.656} & 0.364 & \textbf{0.684}  \\ \cline{2-11} 
 & Schema Matching & DeepM & 1 & 1 & 1 & 1 & 0.857 & \textbf{1} & 1  &  1 \\ \cline{2-11} 
 & \multirow{5}{*}{Row-to-Row Transformation} & BingQL-Unit &  \multicolumn{2}{c||}{\multirow{5}{*}{N.A.}}  & 0.213 & \textbf{0.427} &  \multicolumn{2}{c||}{\multirow{5}{*}{N.A.}}   & 0.339 & \textbf{0.446} \\ \cline{3-3}  \cline{6-7} \cline{10-11}
 &  & BingQL-other &  \multicolumn{2}{c||}{}   & 0.431 & \textbf{0.588}  & \multicolumn{2}{c||}{}  & 0.558 & \textbf{0.607} \\ \cline{3-3}  \cline{6-7}  \cline{10-11}
 &  & FF-GR-Trifacta &  \multicolumn{2}{c||}{}  & 0.712 & \textbf{0.788} & \multicolumn{2}{c||}{}  & 0.772 & \textbf{0.825} \\ \cline{3-3}  \cline{6-7} \cline{10-11}
 &  & Headcase &  \multicolumn{2}{c||}{}  & 0.636 & \textbf{0.705}  & \multicolumn{2}{c||}{}  & 0.704 & \textbf{0.795}  \\ \cline{3-3}  \cline{6-7} \cline{10-11}
 &  & Stackoverflow &  \multicolumn{2}{c||}{}  & 0.662 & \textbf{0.745} & \multicolumn{2}{c||}{}  & \textbf{0.800} & 0.758  \\ \hline
 %\\ \cline{3-3}  \cline{6-7}  
% &  & Wiki &  \multicolumn{2}{c|}{}  & 0.96 & \textbf{0.961} 
% \\ \hline
\end{tabular}
}
\caption{Detailed results of Table-tuning, on both GPT-3.5 and ChatGPT, for individual datasets. Zero-shot is not applicable to row-to-row by-example transformations (marked as ``N.A.''), which requires examples. For all ``Unseen'' tasks, the tasks are held-out and unseen during table-tuning. For all ``Seen'' tasks, the task is seen during table-tuning, but the test datasets are held-out and unseen.}
\end{table*}

\subsection{Experiment Setup}
\stitle{Models Compared.} We test the following models.
\begin{itemize}[leftmargin=*]
    \item \underline{\textit{GPT-3.5 (text-davinci-002)}}. This 175B model is available  from OpenAI, and is one of the vanilla GPT models that we compare with.
    \item \underline{\textit{Table-GPT-3.5 (text-davinci-002 +table-tune)}}. This is the model we obtain by performing table-tuning on GPT-3.5 (text-davinci-002). We compare the performance of Table-GPT-3.5 with GPT-3.5.
    \item \underline{\textit{ChatGPT (text-chat-davinci-002)}}. This is a version of the ChatGPT model available internally~\cite{text-chat-is-chatgpt}, which we use as a second vanilla base model, from which we perform table-tuning.
    \item \underline{\textit{Table-ChatGPT (text-chat-davinci-002 +table-tune)}}. This is the model we obtain by performing table-tuning on ChatGPT (text-chat-davinci-002), which we compare with the vanilla ChatGPT.
\end{itemize}

\stitle{Training tasks and data.} In our default settings, we use a total of 14 types of table-tasks, listed as T-5 to T-18 in Table~\ref{tab:task-summary}, as training data for table-tuning.

In all but two task-types (T-6: Entity Matching and T-12: NL-to-SQL), we use synthesized instances of table-tasks. For each task type, we generate 1000 instances of table-tasks using a 50:50 mix of zero-shot and few-shot templates, following a synthesis-then-augment approach described in Section~\ref{sec:table-tune}.
During task-synthesis, we sample from 125k real web-tables $\mathbf{C}^{wt}$ and database-tables $\mathbf{C}^{db}$ (aggressively deduped from over 2M  original tables).
For Entity Matching and NL-to-SQL where realistic labels/completions cannot be automatically synthesized, we use existing manually-labeled benchmark data, from~\cite{em-data-magellan} and~\cite{nl2sql-spider}, respectively.

%There are two types of sources for data generation:
% \begin{itemize}[leftmargin=*]
%     \item \underline{Real labeled datasets}. We use real labeled benchmark datasets from literature to generate the training examples for Entity Matching~\cite{em-data-magellan} and Natural-Language to SQL~\cite{nl2sql-spider}.
    
%     \item \underline{Synthesized datasets}. We collected 125k tables from Web corpus and PBI corpus, which are then used to synthesize data for the rest of 12 training tasks. Note that for different tasks, we randomly sample tables with no overlapping to increase the diversity.
% \end{itemize}

%For Entity Matching (T-6), Schema Matching (T-7), Data Imputation (T-8), Error Detection (T-9) and Header Value Matching (T-11), we augmented their training data by randomly permute the columns, which doubles the amount of training data for these tasks.

%After removing the training data that exceeds the token limits, the final training data contains 17,909 training examples. 

\stitle{Test tasks and data.} To evaluate the benefit of table-tuning, we test the performance of paired models that are table-tuned vs. vanilla un-tuned, namely, we compare (GPT-3.5 vs. Table-GPT-3.5) and (ChatGPT vs. Table-ChatGPT), as two groups. 

We test on 4 unseen tasks (T-1 to T-4 in Table~\ref{tab:task-summary}), which are completely unseen during table-tuning, to evaluate whether our table-tuned models can continue to follow to human-instructions and perform novel unseen tasks (like illustrated in Figure~\ref{fig:unseen-new-tests}). In addition, we make sure that the test-data used in unseen tasks, are completely separate from the tables used in synthesizing table-tasks as training-data for table-tuning. Specifically, our training data for table-tuning are always drawn from web-tables $\mathbf{C}^{wt}$ and database-tables $\mathbf{C}^{db}$, whereas test-data used in our synthesized table-tasks (T-1: Missing-value identification and T2: Column-finding) are always drawn from a corpus of real spreadsheet tables $\mathbf{C}^{sp}$, completely separate from $\mathbf{C}^{wt}$ and $\mathbf{C}^{db}$ and with very different characteristics. For the remaining two unseen tests (T-3: Table Question and T-4: Column Type Annotation), we use established benchmark data~\cite{table-qa-wikitablequestions} and~\cite{turl, unicorn, cta-sherlock} respectively, which are unseen during table-tuning.

We also evaluate 5 seen tasks (T-5 to T-9 in Table~\ref{tab:task-summary}), which are important table-tasks extensively studied in the literature, which we want table-tuned models to be exposed of to understand these table-related concepts. While these task-types are seen during table-tuning, we make sure that the test datasets are completely separate from the training data used in table-tuning. For synthesized table-tasks (T-8 Data Imputation), similar to discussed above, our test cases are always drawn from a corpus of real spreadsheet tables $\mathbf{C}^{sp}$, separate from the corpus of web-tables $\mathbf{C}^{wt}$ and database-tables $\mathbf{C}^{db}$ used in synthesizing training table-tasks, in order to test table-tuned models' ability to generalize to new tables. For other tasks, we use existing benchmark data, completely unseen when training table-tuned models (e.g.,~\cite{data-transform-tde} for T-5: Row-to-row transformation, \cite{em-data-magellan} for T-6: Entity-matching, using the same setup as~\cite{stanford-prompt-engineer}, \cite{schema-matching-valentine} for T-7: Schema-matching). The task of (T-9) Error-detection is of high value for our business, where we manually labeled a benchmark using real spreadsheet-tables and web-tables for this evaluation.

Details of test data and their statistics can be found in Table~\ref{tab:test_data_details}.

%For Error Detection (T-9), we create two real test datasets by manually labeling errors in real-world spreadsheets and web-tables.  The remaining tasks use existing real benchmark datasets as test data. 

%Table~\ref{tab:test_data_details} shows the detailed statistics of the test tasks and data.

\begin{table}[t]
\caption{Details of test data and evaluation metrics}
\vspace{-3mm}
\label{tab:test_data_details}
\resizebox{\columnwidth}{!}{
\begin{tabular}{|p{1.8in}|c|c|c|}
\hline
\textbf{Task} & \textbf{\begin{tabular}[c]{@{}c@{}}Evaluation \\ Metrics\end{tabular}} & \textbf{Datasets} & \textbf{Size} \\ \hline
\multirow{4}{*}{T-1: Missing Value Identification} & \multirow{4}{*}{F1} & Column (no Separator) & 1000 \\ \cline{3-4} 
 &  & Column (with Separator) & 1000 \\ \cline{3-4} 
 &  & Row (no Separator) & 1000 \\ \cline{3-4} 
 &  & Row (with Separator) & 1000 \\ \hline
T-2: Column Finding & Acc & Spreadsheets-CF & 841 \\ \hline

T-3: Table Question & Acc & Wiki & 4344 \\ \hline

\multirow{4}{*}{T-4: Column Type Annotation} & \multirow{4}{*}{F1} & Efthymiou & 594 \\ \cline{3-4} 
 &  & Limaye & 174 \\ \cline{3-4} 
 &  & Sherlock & 971 \\ \cline{3-4} 
 &  & T2D & 367 \\ \hline

\multirow{5}{*}{T-5: Row-to-Row Transformation} & \multirow{5}{*}{Acc} & BingQL-Unit & 103 \\ \cline{3-4} 
 &  & BingQL-other & 1102 \\ \cline{3-4} 
 &  & FF-GR-Trifacta & 132 \\ \cline{3-4} 
 &  & Headcase & 88 \\ \cline{3-4} 
 &  & Stackoverflow & 145 \\ \hline

\multirow{7}{*}{T-6: Entity Matching} & \multirow{7}{*}{F1} & Amazon-Google & 2293 \\ \cline{3-4} 
 &  & Beer & 91 \\ \cline{3-4} 
 &  & DBLP-ACM & 2473 \\ \cline{3-4} 
 &  & DBLP-GoogleScholar & 5742 \\ \cline{3-4} 
 &  & Fodors-Zagats & 189 \\ \cline{3-4} 
 &  & Walmart-Amazon & 2049 \\ \cline{3-4} 
 &  & iTunes-Amazon & 109 \\ \hline
 
T-7: Schema Matching & Recall & DeepM & 41 \\ \hline
 
T-8: Data Impuation & Acc & Spreadsheets-DI & 1000 \\ \hline

\multirow{2}{*}{T-9: Error Detection} & \multirow{2}{*}{F1} & Spreadsheets-Real & 870 \\ \cline{3-4} 
 &  & WebTables-Real & 432 \\ \hline
\end{tabular}
}
\end{table}

\subsection{Quality Comparisons: Unseen + Seen tasks}

\begin{figure*}[t!]
    \centering    
    \includegraphics[width=0.20\textwidth]{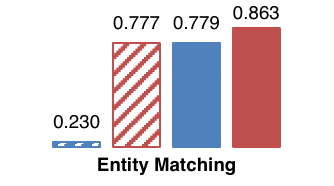}
    \includegraphics[width=0.20\textwidth]{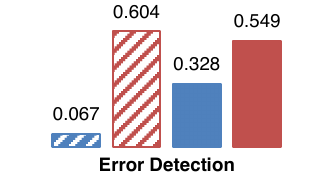}
    \includegraphics[width=0.20\textwidth]{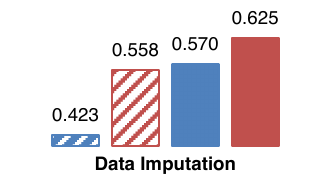}
    \includegraphics[width=0.20\textwidth]{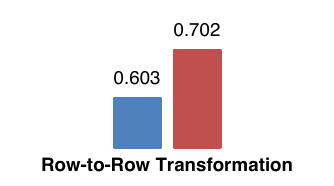}
    \includegraphics[width=0.20\textwidth]{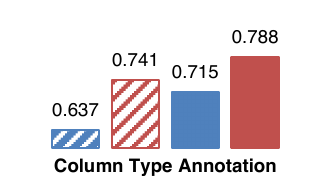}
    \includegraphics[width=0.20\textwidth]{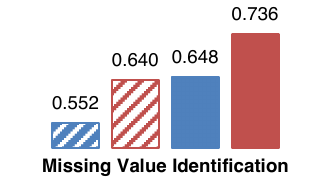}
    \includegraphics[width=0.20\textwidth]{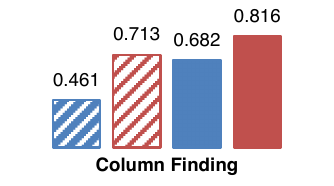}
    \includegraphics[width=0.20\textwidth]{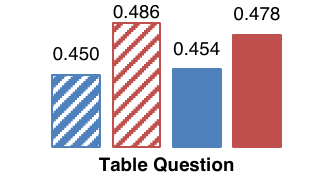}
    \includegraphics[width=0.7\textwidth]{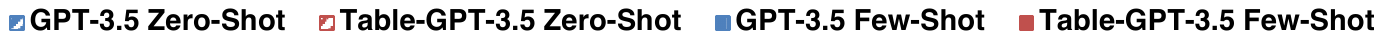}
    \vspace{-2mm}
    \caption{Overall quality improvement, between vanilla GPT-3.5 and Table-GPT-3.5.}
    \label{fig:main-quality-gpt}
\end{figure*}

\begin{figure*}[t!]
    \centering    
    \includegraphics[width=0.20\textwidth]{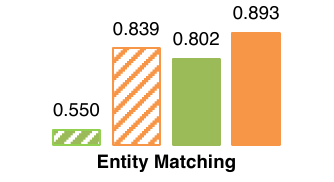}
    \includegraphics[width=0.20\textwidth]{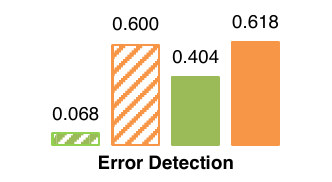}
    \includegraphics[width=0.20\textwidth]{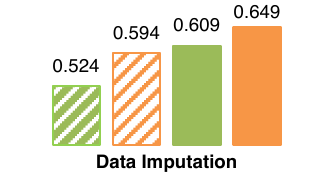}
    \includegraphics[width=0.20\textwidth]{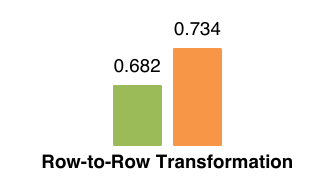}
    \includegraphics[width=0.20\textwidth]{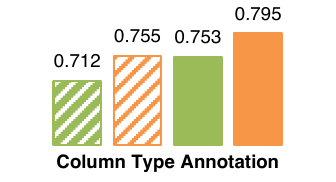}
    \includegraphics[width=0.20\textwidth]{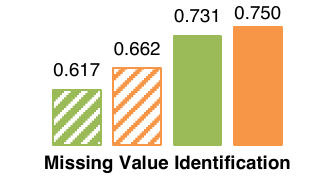}
    \includegraphics[width=0.20\textwidth]{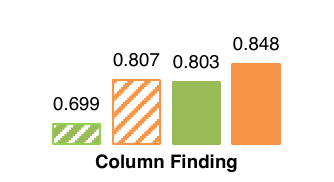}
    \includegraphics[width=0.20\textwidth]{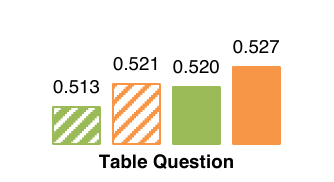}
    \includegraphics[width=0.7\textwidth]{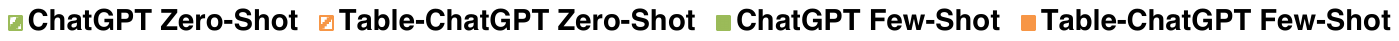}
    \vspace{-2mm}
    \caption{Overall quality improvement, between vanilla ChatGPT and Table-ChatGPT.}
    \label{fig:main-quality-chatgpt}
\end{figure*}

%\stitle{Main Comparison}: 
%Figure~\ref{fig:main-quality-gpt} and Figure~\ref{fig:main-quality-chatgpt} show the main comparisons. 
In Figure~\ref{fig:main-quality-gpt}, we compare the performance between (GPT-3.5 vs. Table-GPT-3.5), and in Figure~\ref{fig:main-quality-chatgpt}, we compare the performance between (ChatGPT vs. Table-ChatGPT), which are table-tuned vs. un-tuned vanilla models, using GPT-3.5 and ChatGPT as base-models, respectively.  Within each task-group in the figures, we show 4 bars, where the first two correspond to zero-shot settings, and the last two correspond to few-shot settings. We can see that across the board, table-tuned models show strong performance benefits on diverse table-tasks. 

It is interesting to note that the benefit of table-tuning is observed when both GPT-3.5 and ChatGPT are used as base-models, showing the generality of our proposed table-tuning approach, on top of underlying language-models of different styles. 

Table~\ref{tab:test_data_details} shows a detailed breakdown of the results, at the individual data-set level. We can see that across 26 test datasets, on 2 base-models (GPT-3.5 and ChatGPT), in 2 settings (zero-shot and few-shot), for a total of 104 tests, table-tuned models outperform their vanilla un-tuned counterparts in 98/104 tests (with the remaining being 3 ties and 3 losses), showing the strong performance benefits of table-tuning. %(when aggregated at the task-type level).

%\stitle{Unseen tasks}: A large table, comparing with Vanilla GPT, Unicorn in detail (e.g.,  diff variants of missing-cells)

%\stitle{Seen tasks}: A large table, compare with Vanilla GPT + various and SOTA, on all seen tasks in detail

%\yeye{leave one out, for seen tasks}

% \textbf{Stanford EM prompt engineering}: focus on EM, compare with Stanford numbers, using Table-GPT + their examples
% 206 brands, amazon-google

% \begin{table}[h]
%     \centering
%     \begin{tabular}{|c|c|c|c|}
%         \hline
%         Row1/Col1 & Row1/Col2 & Row1/Col3 & Row1/Col4 \\
%         \hline
%         Row2/Col1 & \multicolumn{2}{c|}{\multirow{2}{*}{}} & Row2/Col4 \\
%         \cline{1-1} \cline{4-4}
%         Row3/Col1 & \multicolumn{2}{c|}{} & Row3/Col4 \\
%         \hline
%         Row4/Col1 & Row4/Col2 & Row4/Col3 & Row4/Col4 \\
%         \hline
%     \end{tabular}
%     \caption{A table with a 2x2 empty region}
% \end{table}

\subsection{Benefits on task-specific optimizations}
In addition to performing well out-of-the-box in zero-shot and (random) few-shot settings, as shown above, table-tuned GPT models could potentially be used as ``table foundation models'', if they continue to show quality benefits on downstream tasks, when task-specific optimizations are applied. 

Like we discussed in Section~\ref{sec:table-foundation-model}, these include (1) single-task  prompt-engineering, where we select the best instructions and few-shot examples for a single task, using a small number of labeled examples; and (2) single-task fine-tuning, where we continue to fine-tune models for a specific task, with a sufficient number of labeled examples. We will study the benefit of table-tuning in these two settings below.

%To assess the benefit of our table-tuning in enhancing the GPT model's performance on downstream table-tasks, we employ two strategies on both TableGPT and the vanilla model on a single unseen task: (1) task-specific fine-tuning and (2) prompt engineering.

\begin{figure}[h]
    \centering
    \includegraphics[width=\columnwidth]{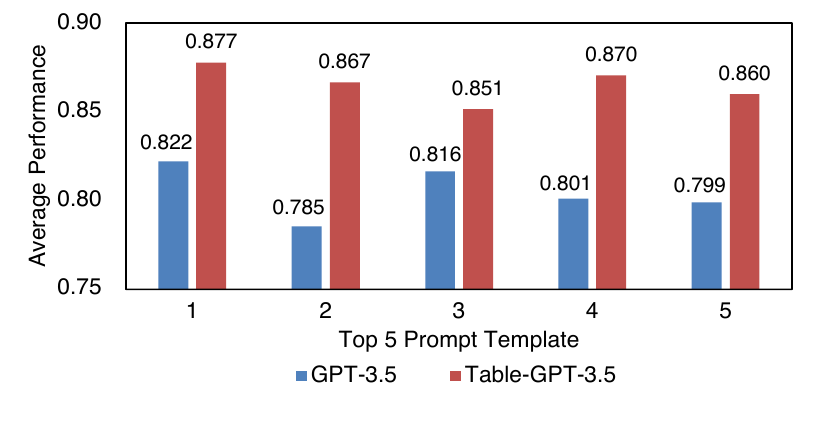}
    \vspace{-6mm}
    \caption{Comparison of quality, when using prompt-engineering. Results shown are for 5 best prompt-templates on the Efthymiou dataset.}
    \label{fig:prompt_engineer}
\end{figure}

\stitle{Single-task prompt-engineering}: We perform prompt-engineering for Table-GPT-3.5 and GPT-3.5, on the column-type-annotation (CTA) task (using the Efthymiou~\cite{turl} dataset), by selecting the best few-shot examples using 200 labeled examples (randomly sampled from the ground-truth), where the goodness of a prompt is evaluated on the labeled examples. %Specifically, we try different prompts by changing the few-shot examples and selecting the best prompt using a validation set. 
Figure~\ref{fig:prompt_engineer} shows the top-5 prompts selected, for Table-GPT-3.5 and GPT-3.5, respectively. We can see that Table-GPT-3.5  consistently outperforms GPT-3.5, on the 5 best prompts produced from prompt-engineering.

\begin{figure}[h]
    \centering
     \begin{subfigure}[t]{0.49\columnwidth}
        \centering
        \includegraphics[width=\columnwidth]{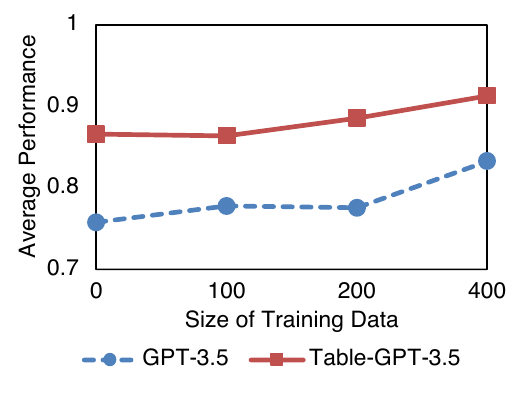}
        \vspace{-6mm}
        \caption{CTA (Efthymiou)}
    \end{subfigure}
     \begin{subfigure}[t]{0.49\columnwidth}
        \centering
        \includegraphics[width=\columnwidth]{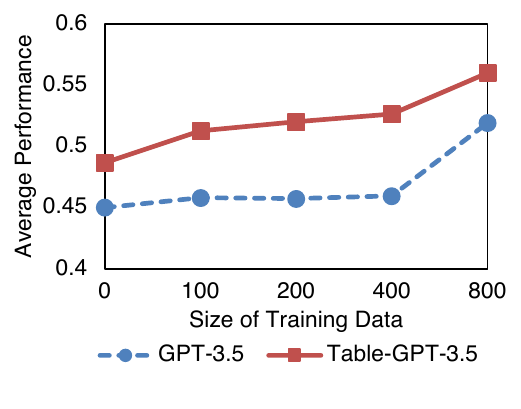}
        \vspace{-6mm}
        \caption{TQA (Wiki)}
    \end{subfigure}
    \vspace{-2mm}
    \caption{Single Task Fine-tuning}
    \label{fig:single_task}
\end{figure}

\stitle{Single-task fine-tuning}: We perform task-specific fine-tuning, on Table-GPT-3.5 and GPT-3.5, using labeled data for that specific task. Table~\ref{fig:single_task}(a) and Table~\ref{fig:single_task}(b) show the comparison, on the CTA task (using  Efthymiou~\cite{turl}) and Table-Question-Answering or TQA (using WikiTableQuestions~\cite{table-qa-wikitablequestions}), respectively. In both cases, we vary the amount of training data on the x-axis. As expected,  the performance of both Table-GPT-3.5 and GPT-3.5 benefit from continued task-specific fine-tuning, but with the same amount of training data, Table-GPT-3.5 continues to dominate GPT-3.5. Looking at the graph from a different way, to achieve the same performance (y-axis), fine-tuning Table-GPT-3.5 would require a smaller number of labeled data than fine-tuning the vanilla GPT-3.5. 

% continue to tune CTA, compare with vanilla GPT and baselines reported in DoDuo paper.

%\yeye{Test prompt tuning exp? cta, select 100 prompt examples, 100 validation examples, randomly pick 30 x 6-example combination, use the best-performing on validation for final test (30x (1000 test + 100 validation)   for Efthymiou is not bad)}

\subsection{Sensitivity Analysis}
We perform sensitivity analysis to better understand table-tuning.

\begin{figure}[h]
\includegraphics[width=\columnwidth]{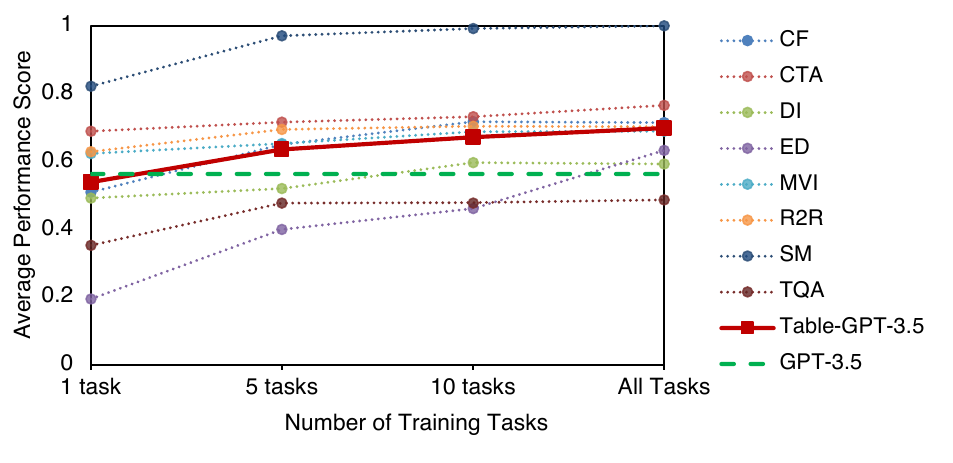}
\vspace{-8mm}
\caption{Vary number of training tasks %(\textcolor{red}{EM is missing})
}
\label{fig:multi_task}
\end{figure}

\stitle{Varying the number of training tasks.} To see whether using more training tasks brings a general benefit, we sample 1/5/10 tasks from all of our training table-tasks for 4 times each, perform fine-tuning on each subset of tasks selected, and compute the average from these runs. 

\begin{figure*}[t]
\centering
\begin{minipage}{.33\textwidth}
    \includegraphics[width=\columnwidth]{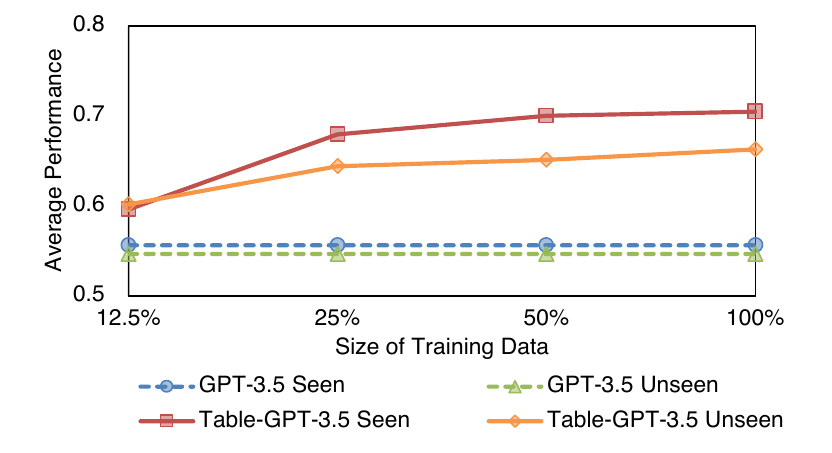}
    \vspace{-8mm}
    \caption{Vary Training Size %(\textcolor{red}{EM is missing})
    }
    \label{fig:vary_train}
\end{minipage}%
\begin{minipage}{.33\textwidth}
    \includegraphics[width=\columnwidth]{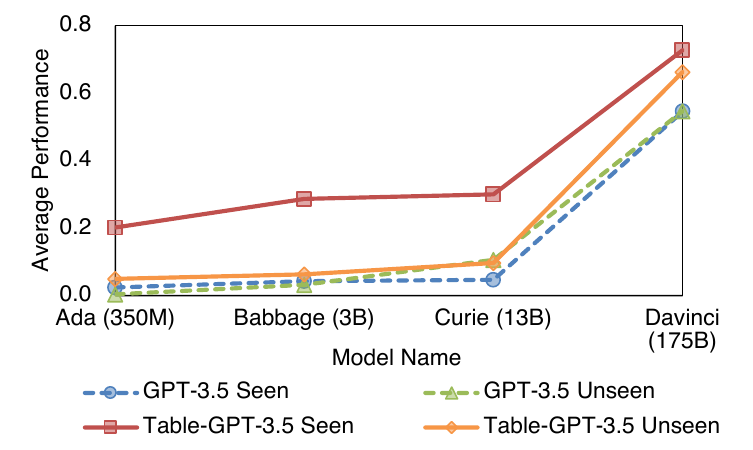}
    \vspace{-8mm}
    \caption{Vary Model Size
    }
    \label{fig:sensitivity_model_size}
\end{minipage}%
\begin{minipage}{.33\textwidth}
    \includegraphics[width=\columnwidth]{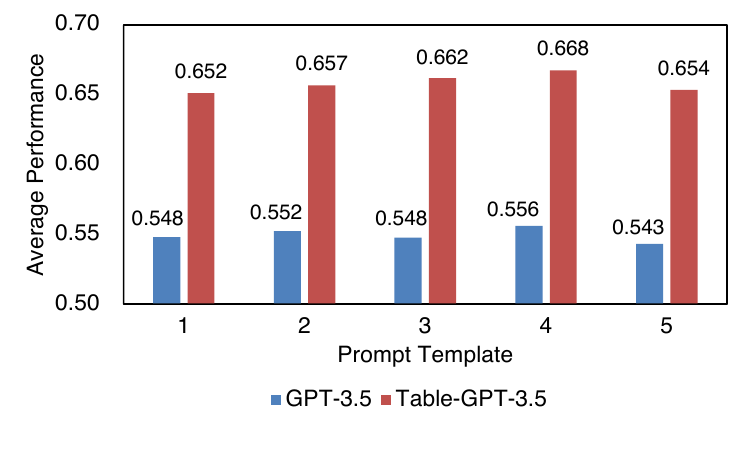}
    \vspace{-8mm}
    \caption{Vary Templates
    }
    \label{fig:sensitivity_template}
\end{minipage}
\end{figure*}

The average quality results are shown in Figure~\ref{fig:vary_train}. As we can see, on the left of the figure with a small number of tasks (e.g., 1), table-tuning degenerates to single-task tuning, which actually hurts the performance of other tasks in general (notice that the performance corresponding to 1-task is lower than the dotted green line, which corresponds to GPT-3.5). As we have more training-tasks, the  performance goes up consistently, for all tasks as well as for the average across all tasks, showing the benefit that is analogous to multi-task training.

\stitle{Vary the amount of training data.} Figure~\ref{fig:vary_train} shows the average performance on seen/unseen test tasks with different amounts of training data. As we can see, the performance on both seen and unseen tasks improves with training data, which plateaus as more data is used. %especially from 12.5\% training to 50\% training. From 50\% to 100\%, the growth of the performance becomes steady.

%\textcolor{blue}{- too much FT, unseen tasks suffer?}

\stitle{Vary base-model Size.} To understand how the size of the base-models affects the performance of table-tuned models, we table-tune four variants of GPT, namely, Text-Ada-001 (350M parameters), Text-Babbage-001 (3B parameters), Text-Curie-001 (13B parameters), Text-Davinci-002 (175B
parameters). 

Figure~\ref{fig:sensitivity_model_size} shows the average performance of base-models vs. table-tuned models, on seen/unseen tasks. We can see that for the unseen tasks (important to check for model generalizability), table-tuned models produce little benefit on smaller models (Ada/Babbage/Curie), but the benefit becomes suddenly significant on larger models (GPT-3.5 and ChatGPT), which appear to be an emerging ability consistent with what is reported in other contexts (e.g.,~\cite{flan, llm-gpt-3}).
%fine-tuned model consistently outperforms the vanilla model across all models on seen/unseen table tasks. Also, we can see that both the performance of TableGPT and the vanilla model grows suddenly from model C to model D, which indicates the emerging behavior of large language models.

% \textbf{Mixed FT with zero-shot and few-shot}
% - diff mixing percentage, mixing helps

\stitle{Vary prompt templates.} To test the robustness of our table-tuned models, we generate 5 different prompt templates (task descriptions and special markers), which are paraphrased automatically using GPT, from a canonical prompt template written by humans. 

Figure~\ref{fig:sensitivity_template}
 shows the average model performance over all unseen test tasks for each prompt template. As we can see, different prompt templates introduce variations in performance, for both Table-GPT-3.5 and GPT-3.5, but the former consistently outperforms the latter by more than 10 percentage points on all 5 prompt templates, showing the robustness of Table-GPT to different kinds of prompts.
 
\stitle{Vary table formats.} There are multiple options when serializing a table $T$ into text, such as Markdown, CSV, JSON, etc. We use the Markdown table format, because it is succinct, and furthermore vanilla GPT tends to generate tables in the Markdown format in responding to human questions, suggesting that it is the table format of its choice, likely because GPT is pre-trained on lots of GitHub code, where Markdown  tables are abundant.  To understand the effect of using different table formats in representing tables in prompts, we test two different table formats, namely CSV and JSON. 

Table~\ref{tab:vary_format} shows the average performance %on seen/unseen/overall tasks 
when using different table formats. As we can see, the Markdown format on average performs better than other formats, although the gap is not too significant.

\begin{table}[h]
\caption{Performance of Table-GPT-3.5, when different table formats are used to serialize tables}
\vspace{-4mm}
\label{tab:vary_format}
\scalebox{0.8}{
\begin{tabular}{c|ccc}
\toprule
\textbf{Task Type} & \textbf{Markdown} & \textbf{CSV} & \textbf{JSON} \\
\midrule
Seen & 0.739 & 0.707 & 0.713 \\
Unseen & 0.663 & 0.662 & 0.621 \\
Overall & 0.705 & 0.687 & 0.672 \\
\bottomrule
\end{tabular}
}
\end{table}

\subsection{Ablation Studies}
We perform ablation analysis to understand the benefit of different augmentation strategies (Section~\ref{sec:augment-tasks}). The results are summarized in Table~\ref{tab:ablation_study}.

%\yeye{no zero-shot / few-shot in ft?}
\begin{table}[h]
\centering
\caption{Ablation Studies of table-tuning}
\vspace{-4mm}
\label{tab:ablation_study}
\resizebox{\columnwidth}{!}{
\begin{tabular}{c|cc|cccc}
\toprule
\textbf{Task Type} & \textbf{GPT-3.5} & \textbf{Table-GPT-3.5} & \textbf{NoSyn} & \textbf{NoColPer.} & \textbf{NoPromptVar.} & \textbf{NoCOT} \\
\midrule
Seen & 0.548 & 0.739 & 0.610 & 0.735 & 0.722 & 0.728 \\
Unseen & 0.547 & 0.663 & 0.607 & 0.661 & 0.657 & 0.666 \\
Overall & 0.548 & 0.705 & 0.608 & 0.702 & 0.693 & 0.701  \\
\bottomrule
\end{tabular}
}
\end{table}

\stitle{No task-level augmentation (no synthesized tasks).} Because we synthesized diverse table-tasks for table-tuning (Section~\ref{sec:synthesize-tasks}), our first ablation is to remove all such tasks from the training data. 
The result is shown in Table~\ref{tab:ablation_study} as ``NoSyn". As we can see, the average performance on seen and unseen tasks drops significantly, showing the contribution of our diverse synthesized table-tasks.

\stitle{No table-level augmentation (no column permutations).} We disable the table-level augmentation by turning off the column permutation. The result is shown in Table~\ref{tab:ablation_study} as ``NoColPer". We can see that the average performance on seen and unseen tasks is lowered without column permutations.

\stitle{No instruction-level augmentation (no prompt variations).} We then disable the instruction-level augmentation, by using only one canonical prompt template for each task (without paraphrasing). The result is shown in Table~\ref{tab:ablation_study} as ``NoPromptVar". As we can see, the average performance of seen and unseen tasks drops slightly, likely because diverse types of table-tasks we use can somewhat mitigate the negative effect of using repeated instruction templates.
%is decreased by 0.004 and 0.002 respectively without prompt variations. The drop is not significant because although each task uses only one prompt template, the prompts are still variant across different tasks as they have different task descriptions.

\stitle{No completion-level augmentation (no chain-of-thought).} We drop the augmentation at completion level by removing the chain-of-thought (COT) reasoning from the completion in the fine-tuning data. The result is shown in Table~\ref{tab:ablation_study} as ``NoCOT". The average performance on seen tasks becomes lower with no COT, which is expected.

%% file: Conclusions.tex
\section{Conclusions and Future Work}
In this work, we propose a new paradigm called table-tuning, that can continue to fine-tune the model weights of pre-trained large language-models like GPT-3.5 and ChatGPT, such that the resulting models are better in  understanding tables and performing table tasks, while still being versatile in following diverse human instructions for unseen tasks. %We conduct extensive experiments to table-tune two base-models, GPT-3.5 and Chat-GPT, into their respective, table-tuned versions, to demonstrate the effectiveness of the proposed approach.
Just like how instruction-tuning has turned into a rich and fruitful line of research in the NLP literature, we hope our initial steps in table-tuning can serve as a springboard for others to continue in this path to develop more optimized models for tables and table-related tasks.

%% file: apx-prompt.tex
\onecolumn
\section{Task details}

\begin{lstlisting}[title=Missing Value Identification (Column No Sep) Zero-Shot]
(*\textbf{\underline{Prompt:}}*)
# Task Description: Please check the following table, there is one and exactly one cell in the table that is missing. When you find this missing cell, please point it out using its column name. Return the final result as JSON in the format {"missing_col": "<missing column name>"}.

## Input:
|Project|Team|Req|
|---|---|---|
|A|5|3|
|I|3|2|
|U|2|3|
|2|1|
|I|2|2|

Return the final result as JSON in the format {"missing_col": "<missing column name>"}.
## Output:

(*\textbf{\underline{Completion:}}*)
{"missing_col": "Project"}
\end{lstlisting}
\begin{lstlisting}[title=Missing Value Identification (Column No Sep) Few-Shot]
(*\textbf{\underline{Prompt:}}*)
# Task Description: Please check the following table, there is one and exactly one cell in the table that is missing. When you find this missing cell, please point it out using its column name. Return the final result as JSON in the format {"missing_col": "<missing column name>"}.

## Input:
|Project|Team|Req|
|---|---|---|
|A|4|1|
|I|2|1|
|O|3|3|
|A|1|
|E|4|2|

## Output:
{"missing_col": "Req"}

## Input:
|Project|Team|Req|
|---|---|---|
|I|2|1|
|E|1|3|
|A|1|3|
|1|2|
|E|4|2|

## Output:
{"missing_col": "Project"}

## Input:
|Project|Team|Req|
|---|---|---|
|E|4|2|
|O|2|
|U|5|2|
|I|4|2|
|A|4|2|

## Output:
{"missing_col": "Team"}

## Input:
|Project|Team|Req|
|---|---|---|
|A|5|3|
|I|3|2|
|U|2|3|
|2|1|
|I|2|2|

Return the final result as JSON in the format {"missing_col": "<missing column name>"}.
## Output:

(*\textbf{\underline{Completion:}}*)
{"missing_col": "Project"}
\end{lstlisting}
\begin{lstlisting}[title=Missing Value Identification (Row Sep) Zero-Shot]
(*\textbf{\underline{Prompt:}}*)
# Task Description: Please check the following table, there is one and exactly one cell in the table that is missing. When you find this missing cell, please point it out using the row id shown in the first column. Return the final result as JSON in the format {"row_id": "<row_id of the row with missing cell>"}.

## Input:
|row_id|Project|Team|Req|
|---|---|---|---|
|1|A|5|3|
|2|I|3|2|
|3|U|2|3|
|4||2|1|
|5|I|2|2|

Return the final result as JSON in the format {"row_id": "<row_id of the row with missing cell>"}.
## Output:

(*\textbf{\underline{Completion:}}*)
{"row_id": "4"}
\end{lstlisting}
\begin{lstlisting}[title=Missing Value Identification (Row Sep) Few-Shot]
(*\textbf{\underline{Prompt:}}*)
# Task Description: Please check the following table, there is one and exactly one cell in the table that is missing. When you find this missing cell, please point it out using the row id shown in the first column. Return the final result as JSON in the format {"row_id": "<row_id of the row with missing cell>"}.

## Input:
|row_id|Project|Team|Req|
|---|---|---|---|
|1|A|4|1|
|2|I|2|1|
|3|O|3|3|
|4|A|1||
|5|E|4|2|

## Output:
{"row_id": "4"}

## Input:
|row_id|Project|Team|Req|
|---|---|---|---|
|1|I|2|1|
|2|E|1|3|
|3|A|1|3|
|4||1|2|
|5|E|4|2|

## Output:
{"row_id": "4"}

## Input:
|row_id|Project|Team|Req|
|---|---|---|---|
|1|E|4|2|
|2|O||2|
|3|U|5|2|
|4|I|4|2|
|5|A|4|2|

## Output:
{"row_id": "2"}

## Input:
|row_id|Project|Team|Req|
|---|---|---|---|
|1|A|5|3|
|2|I|3|2|
|3|U|2|3|
|4||2|1|
|5|I|2|2|

Return the final result as JSON in the format {"row_id": "<row_id of the row with missing cell>"}.
## Output:

(*\textbf{\underline{Completion:}}*)
{"row_id": "4"}
\end{lstlisting}
\begin{lstlisting}[title=Missing Value Identification (Row No Sep) Zero-Shot]
(*\textbf{\underline{Prompt:}}*)
# Task Description: Please check the following table, there is one and exactly one cell in the table that is missing. When you find this missing cell, please point it out using the row id shown in the first column. Return the final result as JSON in the format {"row_id": "<row_id of the row with missing cell>"}.

## Input:
|row_id|Project|Team|Req|
|---|---|---|---|
|1|A|5|3|
|2|I|3|2|
|3|U|2|3|
|4|2|1|
|5|I|2|2|

Return the final result as JSON in the format {"row_id": "<row_id of the row with missing cell>"}.
## Output:

(*\textbf{\underline{Completion:}}*)
{"row_id": "4"}
\end{lstlisting}
\begin{lstlisting}[title=Missing Value Identification (Row No Sep) Few-Shot]
(*\textbf{\underline{Prompt:}}*)
# Task Description: Please check the following table, there is one and exactly one cell in the table that is missing. When you find this missing cell, please point it out using the row id shown in the first column. Return the final result as JSON in the format {"row_id": "<row_id of the row with missing cell>"}.

## Input:
|row_id|Project|Team|Req|
|---|---|---|---|
|1|A|4|1|
|2|I|2|1|
|3|O|3|3|
|4|A|1|
|5|E|4|2|

## Output:
{"row_id": "4"}

## Input:
|row_id|Project|Team|Req|
|---|---|---|---|
|1|I|2|1|
|2|E|1|3|
|3|A|1|3|
|4|1|2|
|5|E|4|2|

## Output:
{"row_id": "4"}

## Input:
|row_id|Project|Team|Req|
|---|---|---|---|
|1|E|4|2|
|2|O|2|
|3|U|5|2|
|4|I|4|2|
|5|A|4|2|

## Output:
{"row_id": "2"}

## Input:
|row_id|Project|Team|Req|
|---|---|---|---|
|1|A|5|3|
|2|I|3|2|
|3|U|2|3|
|4|2|1|
|5|I|2|2|

Return the final result as JSON in the format {"row_id": "<row_id of the row with missing cell>"}.
## Output:

(*\textbf{\underline{Completion:}}*)
{"row_id": "4"}
\end{lstlisting}
\begin{lstlisting}[title=Missing Value Identification (Column Sep) Zero-Shot]
(*\textbf{\underline{Prompt:}}*)
# Task Description: Please check the following table, there is one and exactly one cell in the table that is missing. When you find this missing cell, please point it out using its column name. Return the final result as JSON in the format {"missing_col": "<missing column name>"}.

## Input:
|Project|Team|Req|
|---|---|---|
|A|5|3|
|I|3|2|
|U|2|3|
||2|1|
|I|2|2|

Return the final result as JSON in the format {"missing_col": "<missing column name>"}.
## Output:

(*\textbf{\underline{Completion:}}*)
{"missing_col": "Project"}
\end{lstlisting}
\begin{lstlisting}[title=Missing Value Identification (Column Sep) Few-Shot]
(*\textbf{\underline{Prompt:}}*)
# Task Description: Please check the following table, there is one and exactly one cell in the table that is missing. When you find this missing cell, please point it out using its column name. Return the final result as JSON in the format {"missing_col": "<missing column name>"}.

## Input:
|Project|Team|Req|
|---|---|---|
|A|4|1|
|I|2|1|
|O|3|3|
|A|1||
|E|4|2|

## Output:
{"missing_col": "Req"}

## Input:
|Project|Team|Req|
|---|---|---|
|I|2|1|
|E|1|3|
|A|1|3|
||1|2|
|E|4|2|

## Output:
{"missing_col": "Project"}

## Input:
|Project|Team|Req|
|---|---|---|
|E|4|2|
|O||2|
|U|5|2|
|I|4|2|
|A|4|2|

## Output:
{"missing_col": "Team"}

## Input:
|Project|Team|Req|
|---|---|---|
|A|5|3|
|I|3|2|
|U|2|3|
||2|1|
|I|2|2|

Return the final result as JSON in the format {"missing_col": "<missing column name>"}.
## Output:

(*\textbf{\underline{Completion:}}*)
{"missing_col": "Project"}
\end{lstlisting}

\begin{lstlisting}[title=Column Finding Zero-Shot]
(*\textbf{\underline{Prompt:}}*)
# Task Description: Please look at the table below and find the column that contains the given cell value. Return the final result as JSON in the format {"result": "<name of the column containing the given cell value>"}.

## Input:
**Input Table:**
|1|12|13|14|15|16|17|18|
|---|---|---|---|---|---|---|---|
|2|2|2|2|2|2|2|2|
|3|3|3|3|3|3|3|3|
|4|4|4|4|4|4|4|4|
|5|5|5|5|5|5|5|5|
|6|6|6|6|6|6|6|6|
|7|7|7|7|7|7|7|7|
|8|8|8|8|8|8|8|8|
|9|9|9|9|9|9|9|9|
|10|10|10|10|10|10|10|10|
|11|11|11|11|11|11|11|11|

**Given Cell Value:**
2

Return the final result as JSON in the format {"result": "<name of the column containing the given cell value>"}.
## Output:

(*\textbf{\underline{Completion:}}*)
{"result": "15"}
\end{lstlisting}
\begin{lstlisting}[title=Column Finding Few-Shot]
(*\textbf{\underline{Prompt:}}*)
# Task Description: Please look at the table below and find the column that contains the given cell value. Return the final result as JSON in the format {"result": "<name of the column containing the given cell value>"}.

## Input:
**Input Table:**
|price|crime|nox|rooms|dist|radial|proptax|stratio|
|---|---|---|---|---|---|---|---|
|17794|898299980163574|769999980926514|621000003814697|211999988555908|24|665999984741211|202000007629395|
|21700|384999990463257|769999980926514|63899998664856|250999999046326|24|665999984741211|202000007629395|
|22700|52020001411438|769999980926514|613000011444092|272000002861023|24|665999984741211|202000007629395|
|22600|426100015640259|769999980926514|61100001335144|250999999046326|24|665999984741211|202000007629395|
|24999|454199981689453|769999980926514|640000009536743|251999998092651|24|665999984741211|202000007629395|
|19900|383699989318848|769999980926514|625|229999995231628|24|665999984741211|202000007629395|
|20800|367799997329712|769999980926514|53600001335144|209999990463257|24|665999984741211|202000007629395|
|16800|422200012207031|769999980926514|580000019073486|189999997615814|24|665999984741211|202000007629395|
|21900|347399997711182|717999982833862|877999973297119|189999997615814|24|665999984741211|202000007629395|
|27499|455600023269653|717999982833862|355999994277954|161000001430511|24|665999984741211|202000007629395|

**Given Cell Value:**
426100015640259

## Output:
{"result": "crime"}

## Input:
**Input Table:**
|Player|Class|Team|GP|G|A|Pts|PIM|
|---|---|---|---|---|---|---|---|
|Nathan Gerbe|Junior|Boston College|43|35|33|68|65|
|Kevin Porter|Senior|Michigan|43|33|30|63|18|
|Chad Kolarik|Senior|Michigan|39|30|26|56|24|
|Ryan Lasch|Sophomore|St. Cloud State|40|25|28|53|12|
|Simon Lambert|Senior|RIT|37|21|30|51|40|
|Joe Whitney|Freshman|Boston College|44|11|40|51|50|
|Ben Smith|Sophomore|Boston College|44|25|25|50|12|
|Ryan Jones|Senior|Miami (OH)|42|31|18|49|83|
|Ryan Cruthers|Senior|Robert Morris|34|22|27|49|40|
|Matt Fornataro|Senior|New Hampshire|38|18|28|46|52|

**Given Cell Value:**
Miami (OH)

## Output:
{"result": "Team"}

## Input:
**Input Table:**
|Nation|1977|1995|1997|1999|2001|2003|2005|2008|2010|2012|2014|1979|2016|Years|1981|1983|1985|1987|1989|
|---|---|---|---|---|---|---|---|---|---|---|---|---|---|---|---|---|---|---|---|
|Algeria|nan|nan|nan|nan|nan|19th|nan|16th|nan|nan|nan|nan|nan|4|nan|nan|nan|nan|12th|
|Angola|nan|13th|17th|17th|15th|nan|15th|11th|14th|19th|21st|nan|14th|10|nan|nan|nan|nan|nan|
|Argentina|nan|17th|nan|nan|16th|nan|20th|12th|15th|20th|20th|nan|16th|8|nan|nan|nan|nan|nan|
|Australia|nan|nan|nan|nan|nan|nan|nan|20th|22nd|nan|nan|nan|nan|2|nan|nan|nan|nan|nan|
|Austria|14th|nan|nan|nan|nan|nan|nan|nan|nan|11th|nan|11th|19th|8|nan|nan|14th|nan|14th|
|Belarus|nan|nan|nan|nan|nan|nan|nan|nan|nan|nan|nan|nan|nan|1|nan|nan|nan|nan|nan|
|Brazil|nan|13th|12th|12th|13th|15th|9th|9th|12th|12th|15th|nan|11th|13|nan|nan|nan|nan|nan|
|Bulgaria|nan|9th|nan|nan|nan|nan|nan|nan|nan|nan|nan|nan|nan|5|nan|5th|nan|nan|3rd|
|Canada|nan|nan|nan|20th|nan|nan|nan|nan|nan|nan|nan|nan|nan|3|8th|nan|nan|nan|nan|
|Chile|nan|nan|nan|nan|nan|nan|nan|nan|nan|nan|nan|nan|22nd|1|nan|nan|nan|nan|nan|

**Given Cell Value:**
13

## Output:
{"result": "Years"}

## Input:
**Input Table:**
|1|12|13|14|15|16|17|18|
|---|---|---|---|---|---|---|---|
|2|2|2|2|2|2|2|2|
|3|3|3|3|3|3|3|3|
|4|4|4|4|4|4|4|4|
|5|5|5|5|5|5|5|5|
|6|6|6|6|6|6|6|6|
|7|7|7|7|7|7|7|7|
|8|8|8|8|8|8|8|8|
|9|9|9|9|9|9|9|9|
|10|10|10|10|10|10|10|10|
|11|11|11|11|11|11|11|11|

**Given Cell Value:**
2

Return the final result as JSON in the format {"result": "<name of the column containing the given cell value>"}.
## Output:

(*\textbf{\underline{Completion:}}*)
{"result": "15"}
\end{lstlisting}

\begin{lstlisting}[title=Table-QA Zero-Shot]
(*\textbf{\underline{Prompt:}}*)
# Task Description: Please look at the table, and then answer the question. Please also provide an explanation on your answer. Return the final result as JSON in the format {"answer": "<YOUR ANSWER>"}.

## Input:
*Table*
|Rank|Nation|Gold|Silver|Bronze|Total|
|---|---|---|---|---|---|
|1|Netherlands|8|3|1|12|
|2|Australia|3|3|4|10|
|3|United States|2|5|1|8|
|4|Hungary|1|1|3|5|
|5|Canada|1|-|3|4|
|6|Italy|-|2|1|3|
|7|Russia|-|1|1|2|
|8|China|-|-|1|1|
*Question:*
how many nations are there?

Return the final result as JSON in the format {"answer": "<YOUR ANSWER>"}.
## Output:

(*\textbf{\underline{Completion:}}*)
{"answer": "8"}
\end{lstlisting}
\begin{lstlisting}[title=Table-QA Few-Shot]
(*\textbf{\underline{Prompt:}}*)
# Task Description: Please look at the table, and then answer the question. Please also provide an explanation on your answer. Return the final result as JSON in the format {"answer": "<YOUR ANSWER>"}.

## Input:
*Table*
|Rank|Heat|Nationality|Time|Notes|
|---|---|---|---|---|
|1|1|Russia|23.11|Q|
|2|1|Belgium|23.23|Q|
|3|2|Russia|23.37|Q|
|4|1|Poland|23.39|Q, SB|
|5|1|Ukraine|23.4|Q, SB|
|6|2|United States Virgin Islands|23.49|Q|
|7|1|Belarus|23.52|nan|
|8|1|Canada|23.62|nan|
|9|2|Poland|23.69|Q|
|10|2|Hungary|23.87|Q|
|11|2|Ireland|23.89|nan|
|12|1|Mexico|23.96|nan|
|13|2|Lithuania|24.09|nan|
|14|2|Brazil|24.18|nan|
|15|2|Great Britain|24.31|nan|
|16|1|Italy|24.4|nan|
*Question:*
who is the only cyclist from brazil?

## Output:
{"answer": "Raquel da Costa"}

## Input:
*Table*
|Eps #|Prod #|Title|Air Date|
|---|---|---|---|
|1|2|Menace Of The Mole Men|9/9/1967|
|2|3|Diablo|9/16/1967|
|3|7|The Way It All Began|9/23/1967|
|4|5|Invasion Of The Super-Skrull|9/30/1967|
|5a|1|Klaws|10/7/1967|
|5b|4|The Red Ghost|10/7/1967|
|6|9|Prisoners Of Planet X|10/14/1967|
|7|14|It Started On Yancy Street|10/21/1967|
|8|6|Three Predictions Of Dr. Doom|10/28/1967|
|9|8|Behold A Distant Star|11/4/1967|
|10|12|Demon in the Deep|11/11/1967|
|11|11|Danger In The Depths|11/18/1967|
|12|13|Return Of the Mole Man|11/25/1967|
|13|19|Rama-Tut|12/9/1967|
|14|15|Galactus|12/16/1967|
|15|16|The Micro World Of Dr. Doom|12/30/1967|
|16|17|Blastaar, The Living Bomb-Burst|1/6/1968|
|17|10|The Mysterious Molecule Man|1/13/1968|
|18|18|The Terrible Tribunal|9/14/1968|
|19|20|The Deadly Director|9/21/1968|
*Question:*
how long did the show air?

Return the final result as JSON in the format {"answer": "<YOUR ANSWER>"}.
## Output:

(*\textbf{\underline{Completion:}}*)
{"answer": "1 year"}
\end{lstlisting}

\begin{lstlisting}[title=Column Type Annotation Zero-Shot]
(*\textbf{\underline{Prompt:}}*)
# Task Description: Please look at the input column and determine the semantic type that can describe *every single* instance the input column. Please only choose one semantic type from the candidate list, and remember that the type you choose has to accurately describe every single entity in the column. If no candidate column type can suitably describe every single instance in the column, please return 'None'. Please only choose one type from the candidate list below, and *do not* create new types. Return the final result as JSON in the format {"chosen_semantic_type": "<an entry from the candidate list or None>"}.

## Input:
**Column:**
|Party|
|---|
|Liberal|
|Conservative|

**Candidate column type:**
AcademicJournal
BaseballPlayer
Book
City
Company
Continent
Film
Mayor
Monarch
Mountain
Newspaper
PoliticalParty
Scientist
SportsTeam

Return the final result as JSON in the format {"chosen_semantic_type": "<an entry from the candidate list or None>"}.
## Output:

(*\textbf{\underline{Completion:}}*)
{"chosen_semantic_type": "PoliticalParty"}
\end{lstlisting}
\begin{lstlisting}[title=Column Type Annotation Few-Shot]
(*\textbf{\underline{Prompt:}}*)
# Task Description: Please look at the input column and determine the semantic type that can describe *every single* instance the input column. Please only choose one semantic type from the candidate list, and remember that the type you choose has to accurately describe every single entity in the column. If no candidate column type can suitably describe every single instance in the column, please return 'None'. Please only choose one type from the candidate list below, and *do not* create new types. Return the final result as JSON in the format {"chosen_semantic_type": "<an entry from the candidate list or None>"}.

## Input:
**Column:**
|Name|
|---|
|Wells Fargo Tower|
|Regions-Harbert Plaza|
|AT&T City Center|
|Regions Center|
|City Federal Building|
|Alabama Power Headquarters Building|
|Thomas Jefferson Tower|
|John Hand Building|
|Daniel Building|

**Candidate column type:**
AcademicJournal
Airport
Book
Building
City
Film
Mammal
Newspaper
Plant
PoliticalParty
Scientist
SportsTeam

## Output:
{"chosen_semantic_type": "Building"}

## Input:
**Column:**
|Team|
|---|
|Minnesota 1|
|New York|

**Candidate column type:**
AdministrativeRegion
Continent
Mayor
Saint
University
Wrestler
Writer

## Output:
{"chosen_semantic_type": "AdministrativeRegion"}

## Input:
**Column:**
|Engine|
|---|
|Honda|
|Honda|
|Honda|
|Chevrolet|
|Chevrolet|

**Candidate column type:**
AcademicJournal
Company
Currency
Film
Lake
Saint
Writer

## Output:
{"chosen_semantic_type": "Company"}

## Input:
**Column:**
|Name|
|---|
|Juliomys|
|Pipanacoctomys|
|Salinoctomys|
|Tapecomys|
|Hyladelphys|
|Handleyomys|
|Sommeromys|
|Chacodelphys|
|Drymoreomys|

**Candidate column type:**
Building
City
Company
Continent
Country
Currency

## Output:
{"chosen_semantic_type": "None"}

## Input:
**Column:**
|Service|
|---|
|Star FM|
|Radio Teddy|
|Berliner Rundfunk 91|
|4|
|Jam FM|
|94|
|3 rs2|
|Radio Eins|
|Deutschlandfunk|
|98.8 KISS FM Berlin|
|Antenne Brandenburg|
|FluxFM|
|Klassik Radio|
|Radyo Metropol FM|
|Fritz|
|Energy Berlin|
|104.6 RTL|
|105'5 Spreeradio|
|Radio B2|

**Candidate column type:**
BaseballPlayer
Building
Lake
Mammal
Mayor
PoliticalParty
SportsTeam

## Output:
{"chosen_semantic_type": "None"}

## Input:
**Column:**
|Common name|
|---|
|White-collared swift|

**Candidate column type:**
Airport
Book
Language
Monarch
Plant
PoliticalParty
University

## Output:
{"chosen_semantic_type": "None"}

## Input:
**Column:**
|Party|
|---|
|Liberal|
|Conservative|

**Candidate column type:**
AcademicJournal
BaseballPlayer
Book
City
Company
Continent
Film
Mayor
Monarch
Mountain
Newspaper
PoliticalParty
Scientist
SportsTeam

Return the final result as JSON in the format {"chosen_semantic_type": "<an entry from the candidate list or None>"}.
## Output:

(*\textbf{\underline{Completion:}}*)
{"chosen_semantic_type": "PoliticalParty"}
\end{lstlisting}

\begin{lstlisting}[title=Error Detection Zero-Shot]
(*\textbf{\underline{Prompt:}}*)
Description: Please examine the input table and let me know which cell or cells are erroneous. If there are multiple erroneous cells, return them all in a list. If there is no erroneous cell, return 'None'. Please only return the erroneous cell or cells with high confidence. Return the final result as JSON in the format {"erroneous_cells": "<a list of erroneous cells or None>"}.  Let's think step by step and show your reasoning before showing the final result.

Question:
|Province|Teams|Rank|
|---|---|---|
|Bengkulu|1|9|
|Lampung|1|9|
|Banetn|1|9|
|Central MJava|1|9|
|Yogyakarta|1|9|

Let's think step by step and show your reasoning before showing the final result. Return the final result as JSON in the format {"erroneous_cells": "<a list of erroneous cells or None>"}.
Answer:

(*\textbf{\underline{Completion:}}*)
The cell "Banetn" is misspelled and should be "Banten". The cell "Central MJava" is misspelled and should be "Central Java". Therefore, the answer is {"erroneous_cells": ["Banetn", "Central MJava"]}
\end{lstlisting}
\begin{lstlisting}[title=Error Detection Few-Shot]
(*\textbf{\underline{Prompt:}}*)
# Task Description: Kindly evaluate the input table and let me know about any cell or cells that have errors. If there are multiple cells with errors, provide them in a list. If no cells have errors, indicate 'None'. Only return the cells for which you have a high level of confidence in their error status. Return the final result as JSON in the format {"erroneous_cells": "<a list of erroneous cells or None>"}.  Let's think step by step and show your reasoning before showing the final result.

In:
|District|Party|First Election|
|---|---|---|
|12|Repp|2011|
|43|Dem|2003|
|24|Rep|2002|

Out:
The cell "Repp" is misspelled and should be "Rep". Therefore, the answer is {"erroneous_cells": ["Repp"]}

In:
|District|Party|First Election|
|---|---|---|
|1|Rep|1993|
|2|Rep|2015|
|3|Rep|2009|
|4|Rep|2013|
|5|Rep|2011|

Let's think step by step and show your reasoning before showing the final result. Return the final result as JSON in the format {"erroneous_cells": "<a list of erroneous cells or None>"}.
Out:

(*\textbf{\underline{Completion:}}*)
All cells in the table seem to be correct and no apparent error is found. Therefore, the final answer is {"erroneous_cells": "None"}
\end{lstlisting}
\begin{lstlisting}[title=Row/Column Swapping Zero-Shot]
(*\textbf{\underline{Prompt:}}*)
# Task Description: Please swap the second row and the third row in the table. Please return the table after swapping.

Q:
|Value|Colour|Type|
|---|---|---|
|1 lepton|brown|a|
|2 lepta|grey|a|
|3 lepta|orange|a|
|5 lepta|green|b|
|10 lepta|red|b|

A:

(*\textbf{\underline{Completion:}}*)
|Value|Colour|Type|
|---|---|---|
|1 lepton|brown|a|
|3 lepta|orange|a|
|2 lepta|grey|a|
|5 lepta|green|b|
|10 lepta|red|b|
\end{lstlisting}
\begin{lstlisting}[title=Row/Column Swapping Few-Shot]
(*\textbf{\underline{Prompt:}}*)
Objective: If possible, could you swap the positions of column "Group" and column "%" in the table? Let me know the outcome.

## Input:
|Group|N|%|
|---|---|---|
|IIG|0|0.0%|
|IIIAB|11|1.1%|
|IVB|0|0.0%|
|IIF|1|0.1%|
|IIE|2|0.2%|

## Output:
|%|N|Group|
|---|---|---|
|0.0%|0|IIG|
|1.1%|11|IIIAB|
|0.0%|0|IVB|
|0.1%|1|IIF|
|0.2%|2|IIE|

## Input:
|Group|N|%|
|---|---|---|
|IAB complex|10|1.0%|
|IC|0|0.0%|
|IIAB|6|0.6%|
|IIC|0|0.0%|
|IID|3|0.3%|

## Output:

(*\textbf{\underline{Completion:}}*)
|%|N|Group|
|---|---|---|
|1.0%|10|IAB complex|
|0.0%|0|IC|
|0.6%|6|IIAB|
|0.0%|0|IIC|
|0.3%|3|IID|
\end{lstlisting}
\begin{lstlisting}[title=Schema Matching Zero-Shot]
(*\textbf{\underline{Prompt:}}*)
Description: Please identify the matching columns between Table A and Table B. For each column in Table A, specify the corresponding column in Table B. If a column in A has no corresponding column in Table B, you can map it to None. Represent each column mapping using a pair of column headers in a list, i.e., [Table A Column, Table B column or None]. Provide the mapping for each column in Table A and return all mappings in a list. Return the final result as JSON in the format {"column_mappings": "<a list of column pairs>"}.

Question:
Table A:
|IATA|ICAO|Country|
|---|---|---|
|YYC|CYYC|Canada|
|BNE|YBBN|Australia|

Table B:
|Nation|Airport Code|
|---|---|
|Thailand|DMK|
|Philippines|CEB|
|Philippines|CRK|

Return the final result as JSON in the format {"column_mappings": "<a list of column pairs>"}.
Answer:

(*\textbf{\underline{Completion:}}*)
{"column_mappings": [["IATA", "Airport Code"], ["ICAO", "None"], ["Country", "Nation"]]}
\end{lstlisting}
\begin{lstlisting}[title=Schema Matching Few-Shot]
(*\textbf{\underline{Prompt:}}*)
Description: Find the corresponding columns between Table A and Table B. Specify the related columns in Table B for each column present in Table A. If a column in Table A has no corresponding column in Table B, you can denote it as None. Use pairs of column headers in a list structure, like [Table A Column, Table B Column or None], to depict each mapping. Present the mappings for all the columns in Table A and organize them into a list Return the final result as JSON in the format {"column_mappings": "<a list of column pairs>"}.

Question:
Table A:
|Dignity|Detriment|
|---|---|
|Gemini and Virgo|Sagittarius and Pisces|
|Leo|Aquarius|

Table B:
|Celestial Body (Symbol)|High Point|
|---|---|
|Moon ()|Taurus|
|Venus ()|Pisces|
|Mars ()|Capricorn|

Answer:
{"column_mappings": [["Dignity", "None"], ["Detriment", "None"]]}

Question:
Table A:
|Barony|Civil parish|Acres|Poor law union|County|
|---|---|---|---|---|
|Callan|Callan|21|Callan|Kilkenny|
|Callan|Callan|xx|Callan|Kilkenny|

Table B:
|Location|Parish|Surface Area|Subdivision|
|---|---|---|---|
|Caherlesk|Ballytobin|1,162|Kells|
|CLOGH T.|Castlecomer|xx|Fassadinin|
|CASTLECOMER T.|Castlecomer|xx|Fassadinin|

Return the final result as JSON in the format {"column_mappings": "<a list of column pairs>"}.
Answer:

(*\textbf{\underline{Completion:}}*)
{"column_mappings": [["Barony", "Subdivision"], ["Civil parish", "Parish"], ["Acres", "Surface Area"], ["Poor law union", "None"], ["County", "None"]]}
\end{lstlisting}
\begin{lstlisting}[title=Header Value Matching Zero-Shot]
(*\textbf{\underline{Prompt:}}*)
Task: Given the input table data and the list of candidate headers, please determine the most suitable column header for each column in the table. Please only choose column headers from the candidate list. Please only return the most suitable column header for each column. Return the chosen column headers in a list. Do not return the entire table. Return the final result as JSON in the format {"column_headers": "<a list of headers for each column chosen from the candidate list>"}.

[Q]:
**Table Data:**
||||
|---|---|---|
|1|681|Multihull|
|3|911|Keelboat|
|2|947|Dinghy|
|nan|920|Dinghy|
|1|870|Dinghy|

**Candidate column headers:**
- Portsmouth Number
- crew
- Type

Return the final result as JSON in the format {"column_headers": "<a list of headers for each column chosen from the candidate list>"}.
[A]:

(*\textbf{\underline{Completion:}}*)
{"column_headers": ["crew", "Portsmouth Number", "Type"]}
\end{lstlisting}
\begin{lstlisting}[title=Header Value Matching Few-Shot]
(*\textbf{\underline{Prompt:}}*)
Objective: With the input table data and the list of potential headers at hand, make determinations about the most fitting column header for every table column. Only select column headers from the candidate list and compile them into a list. Return the final result as JSON in the format {"column_headers": "<a list of headers for each column chosen from the candidate list>"}.

## Input:
**Table Data:**
||||
|---|---|---|
|Cust 1|126|1|
|Cust 2|65|2|
|Cust 5|30|5|
|Cust 3|45|3|
|Cust 9|19|9|

**Candidate column headers:**
- Count
- Custom
- Name

## Output:
{"column_headers": ["Name", "Count", "Custom"]}

## Input:
**Table Data:**
||||
|---|---|---|
|1|The Beginning|1:48|
|15|Ride Slow|3:05|
|7|Siri (Skit)|0:53|
|5|Wrong n Tough|2:45|
|9|Officer Hatchet|3:46|

**Candidate column headers:**
- Name
- Length
- Track #

Return the final result as JSON in the format {"column_headers": "<a list of headers for each column chosen from the candidate list>"}.
## Output:

(*\textbf{\underline{Completion:}}*)
{"column_headers": ["Track #", "Name", "Length"]}
\end{lstlisting}
\begin{lstlisting}[title=Column Augmentation Zero-Shot]
(*\textbf{\underline{Prompt:}}*)
Task: Please generate a new additional column for the input table and append the new column to the right of the table. Return the new table with the additional column.

Input:
|Number|Name|
|---|---|
|279|Hazel|
|280|Audrey|
|281|Gwen|
|282|Doris|
|283|Mona|
|284|Vera|
|285|Car No 85|
|286|Car No 86|
|287|Car No 87|
|288|Car No 88|

Output:

(*\textbf{\underline{Completion:}}*)
|Number|Name|Note|
|---|---|---|
|279|Hazel|5-BEL Trust, under restoration|
|280|Audrey|VSOE, in service|
|281|Gwen|VSOE, in service|
|282|Doris|5-BEL Trust, to be restored|
|283|Mona|VSOE, stored|
|284|Vera|VSOE, in service|
|285|Car No 85|5BEL Trust, under restoration|
|286|Car No 86|VSOE, stored|
|287|Car No 87|5-BEL Trust, under restoration|
|288|Car No 88|5-BEL Trust, under restoration|
\end{lstlisting}
\begin{lstlisting}[title=Column Augmentation Few-Shot]
(*\textbf{\underline{Prompt:}}*)
Task: Generate a new column for the input table and attach it on the right side. Share the updated table with the appended column.

## Input:
|Year|Peak positions(CAN Country)|
|---|---|
|1981|41|
|1983|40|
|1983|35|
|1984|39|
|1985|29|
|1986|10|
|1987|20|
|1988|9|
|1989|30|
|1989|9|

## Output:
|Year|Peak positions(CAN Country)|Album|
|---|---|---|
|1981|41|Non-album songs|
|1983|40|Non-album songs|
|1983|35|Non-album songs|
|1984|39|Touch My Heart|
|1985|29|Touch My Heart|
|1986|10|Touch My Heart|
|1987|20|Touch My Heart|
|1988|9|Touch My Heart|
|1989|30|Touch My Heart|
|1989|9|Touch My Heart|

## Input:
|end km|home km|
|---|---|
|0.8|1.402|
|1.402|1.44|
|1.44|1.884|
|1.884|6.071|
|6.071|6.513|
|6.513|8.3|
|8.3|12.4|
|12.4|14.6|
|14.6|21.0|
|21.0|168.7|

## Output:

(*\textbf{\underline{Completion:}}*)
|end km|home km|km/h|
|---|---|---|
|0.8|1.402|110|
|1.402|1.44|35|
|1.44|1.884|65|
|1.884|6.071|110|
|6.071|6.513|105|
|6.513|8.3|110|
|8.3|12.4|165|
|12.4|14.6|200|
|14.6|21.0|270|
|21.0|168.7|300|
\end{lstlisting}
\begin{lstlisting}[title=Natural-Language to SQL Zero-Shot]
(*\textbf{\underline{Prompt:}}*)
Description: Please write the SQL query given the input table and question. Use 'table' as the table name if needed. Return the final result as JSON in the format {"SQL": "<SQL code>"}.

In:
**Input table:**
|Rank|Group|Name|Nationality|Mark|
|---|---|---|---|---|
|1|B|Brittney Reese|United States|6.87|
|2|A|Maurren Higa Maggi|Brazil|6.79|
|3|B|Tatyana Lebedeva|Russia|6.70|
|4|A|Carolina Kluft|Sweden|6.70|
|5|A|Grace Upshaw|United States|6.68|

**Question:**
What rank does Tricia Flores have?

Return the final result as JSON in the format {"SQL": "<SQL code>"}.
Out:

(*\textbf{\underline{Completion:}}*)
{"SQL": "SELECT  `Rank` FROM table WHERE `Name` = tricia flores"}
\end{lstlisting}
\begin{lstlisting}[title=Natural-Language to SQL Few-Shot]
(*\textbf{\underline{Prompt:}}*)
Description: Please write an SQL query using the input table and question. If needed, utilize 'table' as the table name. Return the final result as JSON in the format {"SQL": "<SQL code>"}.

Q:
**Input table:**
|Date|Score|Set 1|Set 2|Set 3|Total|
|---|---|---|---|---|---|
|13 Nov|3-0|15-6|15-12|15-13|45-31|
|13 Nov|0-3|6-15|13-15|6-15|25-45|
|14 Nov|0-3|2-15|8-15|6-15|16-45|
|14 Nov|3-2|15-11|10-15|16-14|67-64|
|15 Nov|3-1|15-9|15-11|13-15|58-47|

**Question:**
What is the score for set 2 when set 3 was 15-6?

A:
{"SQL": "SELECT  `Set 2` FROM table WHERE `Set 3` = 15-6"}

Q:
**Input table:**
|Position|Team|Played|Wins|Draws|Losses|Scored|Conceded|Points|
|---|---|---|---|---|---|---|---|---|
|1|Olimpia|9|4|4|1|14|6|16|
|2|12 de Octubre|9|3|6|0|14|8|15|
|3|Cerro Porteno|9|3|5|1|16|10|14|
|4|Guarani|9|3|5|1|8|6|14|
|5|Cerro Cora|9|3|4|2|9|8|13|

**Question:**
How many losses for the team atl. colegiales?

Return the final result as JSON in the format {"SQL": "<SQL code>"}.
A:

(*\textbf{\underline{Completion:}}*)
{"SQL": "SELECT MAX `Losses` FROM table WHERE `Team` = Atl. Colegiales"}
\end{lstlisting}
\begin{lstlisting}[title=Table Summary Zero-Shot]
(*\textbf{\underline{Prompt:}}*)
Objective: Please look at the table below and provide a title for the table. Return the final result as JSON in the format {"summary": "<summary of table>"}.

## Input:
|Unnamed: 0|Day 8|Nominations received|Day 9|Day 10|Day 13|
|---|---|---|---|---|---|
|Jenny|Lucy|1.0|Georgina|Lucy|Natalia|
|Natalia|In Secret Loft|13.0|Georgina|Lucy|Manuel|
|Marijke|Jan|18.0|Jan|Lucy|Fancy|
|Simon|Marijke|5.0|Marijke|Marijke|Marijke|
|Martin|Lucy|3.0|Lucy|Lucy|Natalia|

Return the final result as JSON in the format {"summary": "<summary of table>"}.
## Output:

(*\textbf{\underline{Completion:}}*)
{"summary": "Promi Big Brother (season 1)"}
\end{lstlisting}
\begin{lstlisting}[title=Table Summary Few-Shot]
(*\textbf{\underline{Prompt:}}*)
Objective: Your task is to give a summary of the input table's main information. Return the final result as JSON in the format {"summary": "<summary of table>"}.

In:
|Survivor|Day 1|Day 2|Day 3|Day 4|Day 5|Day 6|Day 7|Missions|
|---|---|---|---|---|---|---|---|---|
|Amena|SAFE|SAFE|SAFE|SAFE|SAFE|SAFE|WON|7|
|Leah|SAFE|SAFE|SAFE|SAFE|SAFE|SAFE|WON|6|
|Megan|SAFE|SAFE|SAFE|SAFE|SAFE|SAFE|WON|7|
|Nic|NOT IN GAME|NOT IN GAME|NOT IN GAME|SAFE|SAFE|SAFE|WON|5|
|Aston|SAFE|SAFE|SAFE|SAFE|SAFE|SAFE|DIED|7|

Out:
{"summary": "I Survived a Zombie Apocalypse"}

In:
|Name|Charter/Plan|Type|Class|County|
|---|---|---|---|---|
|Bensalem|Plan|Township|Second|Bucks|
|Bethlehem|Charter|City|Third|Lehigh Northampton|
|Bristol|Plan|Township|First|Bucks|
|College|Plan|Township|Second|Centre|
|DuBois|Plan|City|Third|Clearfield|

Return the final result as JSON in the format {"summary": "<summary of table>"}.
Out:

(*\textbf{\underline{Completion:}}*)
{"summary": "List of Pennsylvania municipalities and counties with home rule charters, optional charters, or optional plans"}
\end{lstlisting}
\begin{lstlisting}[title=List Extraction Zero-Shot]
(*\textbf{\underline{Prompt:}}*)
Objective: Please transform the list below into a table with multiple columns. Please return the table using plain text. Use vertical bars (|) to separate columns and use a new line for each row. Return the final result as JSON in the format {"table": "<table transformed from the list>"}.

[Q]:
JobFamily OrgID minTiG maxTiG
JF1 5.0 2.0 12.0
JF3 1.0 2.0 16.0
JF2 4.0 2.0 7.0
JF1 9.0 2.0 7.0
JF3 41.0 2.0 7.0

Return the final result as JSON in the format {"table": "<table transformed from the list>"}.
[A]:

(*\textbf{\underline{Completion:}}*)
{"table": "|JobFamily|OrgID|minTiG|maxTiG|\n|---|---|---|---|\n|JF1|5.0|2.0|12.0|\n|JF3|1.0|2.0|16.0|\n|JF2|4.0|2.0|7.0|\n|JF1|9.0|2.0|7.0|\n|JF3|41.0|2.0|7.0|\n"}
\end{lstlisting}
\begin{lstlisting}[title=List Extraction Few-Shot]
(*\textbf{\underline{Prompt:}}*)
Description: Change the list provided into a table with several columns. The table must be represented in plain text, with vertical bars (|) separating columns and a new line for every row. Return the final result as JSON in the format {"table": "<table transformed from the list>"}.

[Q]:
Ch Video (MHz) Audio (MHz)
15 487.25 493.75
36 693.25 699.75
39 717.25 723.75
23 551.25 557.75
58 871.25 877.75

[A]:
{"table": "|Ch|Video (MHz)|Audio (MHz)|\n|---|---|---|\n|15|487.25|493.75|\n|36|693.25|699.75|\n|39|717.25|723.75|\n|23|551.25|557.75|\n|58|871.25|877.75|\n"}

[Q]:
Month name amount
5 c 3
4 e 2
3 b 6
2 e 2
3 a 3

Return the final result as JSON in the format {"table": "<table transformed from the list>"}.
[A]:

(*\textbf{\underline{Completion:}}*)
{"table": "|Month|name|amount|\n|---|---|---|\n|5|c|3|\n|4|e|2|\n|3|b|6|\n|2|e|2|\n|3|a|3|\n"}
\end{lstlisting}
\begin{lstlisting}[title=Row Augmentation Zero-Shot]
(*\textbf{\underline{Prompt:}}*)
Task: Please generate a new row for the input table and append it at the bottom of the table. Return the new table with the additional generated row.

[Q]:
|Rank|Player|Receptions|
|---|---|---|
|1|Jerry Rice|1,549|
|2|Tony Gonzalez|1,325|
|3|Larry Fitzgerald|1,125|
|4|Marvin Harrison|1,102|
|5|Cris Carter|1,101|
|6|Tim Brown|1,094|
|7|Jason Witten|1,089|
|8|Terrell Owens|1,078|
|9|Anquan Boldin|1,076|

[A]:

(*\textbf{\underline{Completion:}}*)
|Rank|Player|Receptions|
|---|---|---|
|1|Jerry Rice|1,549|
|2|Tony Gonzalez|1,325|
|3|Larry Fitzgerald|1,125|
|4|Marvin Harrison|1,102|
|5|Cris Carter|1,101|
|6|Tim Brown|1,094|
|7|Jason Witten|1,089|
|8|Terrell Owens|1,078|
|9|Anquan Boldin|1,076|
|10|Reggie Wayne|1,070|
\end{lstlisting}
\begin{lstlisting}[title=Row Augmentation Few-Shot]
(*\textbf{\underline{Prompt:}}*)
Description: Please generate a new row for the input table and append it at the bottom of the table. Return the new table with the additional generated row.

[Q]:
|ProductID|Month|TargetSale|
|---|---|---|
|1|6|50|
|1|7|60|
|1|8|70|
|2|6|40|
|2|7|50|
|2|8|60|
|3|6|30|
|3|7|40|
|3|8|50|

[A]:
|ProductID|Month|TargetSale|
|---|---|---|
|1|6|50|
|1|7|60|
|1|8|70|
|2|6|40|
|2|7|50|
|2|8|60|
|3|6|30|
|3|7|40|
|3|8|50|
|4|6|30|

[Q]:
|Rank|Player|2015 FF Points|Games|Avg|
|---|---|---|---|---|
|1|Cam Newton|389.1|16|24.3|
|2|Tom Brady|343.7|16|21.5|
|3|Russell Wilson|336.4|16|21.0|
|4|Blake Bortles|316.1|16|19.8|
|5|Carson Palmer|309.2|16|19.3|
|7|Aaron Rodgers|301.3|16|18.8|
|8|Kirk Cousins|293.5|16|18.3|
|9|Matthew Stafford|289.7|16|18.1|
|10|Eli Manning|287.6|16|18.0|

[A]:

(*\textbf{\underline{Completion:}}*)
|Rank|Player|2015 FF Points|Games|Avg|
|---|---|---|---|---|
|1|Cam Newton|389.1|16|24.3|
|2|Tom Brady|343.7|16|21.5|
|3|Russell Wilson|336.4|16|21.0|
|4|Blake Bortles|316.1|16|19.8|
|5|Carson Palmer|309.2|16|19.3|
|7|Aaron Rodgers|301.3|16|18.8|
|8|Kirk Cousins|293.5|16|18.3|
|9|Matthew Stafford|289.7|16|18.1|
|10|Eli Manning|287.6|16|18.0|
|11|Ryan Fitzpatrick|285.1|16|17.8|
\end{lstlisting}
\begin{lstlisting}[title=Row/Column Filtering Zero-Shot]
(*\textbf{\underline{Prompt:}}*)
# Task Description: Please select the first row in the table. Please return the table with the selected rows.

Question:
|sk_usuario|nk_id_usuario|nome_usuario|
|---|---|---|
|1|1|tatsurok2018|
|4|4|Wilfredo M|
|12|12|Jonathan C|
|19|19|kenthecelt|
|20|20|josenc2016|

Answer:

(*\textbf{\underline{Completion:}}*)
|sk_usuario|nk_id_usuario|nome_usuario|
|---|---|---|
|1|1|tatsurok2018|
\end{lstlisting}
\begin{lstlisting}[title=Row/Column Filtering Few-Shot]
(*\textbf{\underline{Prompt:}}*)
Objective: Could you choose column "Option_content" in the table? After the selection, please return the table.

In:
|question_id|Ans_id|Option_content|
|---|---|---|
|94|2|False|
|111|1|False|
|118|2|False|
|57|1|False|
|95|1|False|

Out:
|Option_content|
|---|
|False|
|False|
|False|
|False|
|False|

In:
|question_id|Ans_id|Option_content|
|---|---|---|
|16|1|False|
|17|1|False|
|18|1|False|
|19|2|False|
|20|2|False|

Out:

(*\textbf{\underline{Completion:}}*)
|Option_content|
|---|
|False|
|False|
|False|
|False|
|False|
\end{lstlisting}
\begin{lstlisting}[title=Row-to-Row Transformation Zero-Shot]
(*\textbf{\underline{Prompt:}}*)
Task: You are given a table of inputs and outputs in two columns. Please figure out the patterns between inputs and outputs from the first few rows and then determine the output value for the last row (denoted by '[Output Value]'). Please only return the output value. Return the final result as JSON in the format {"output_value": "<output value transformed from the input value>"}.

[Q]:
|Input|Output|
|---|---|
|18:1|1|
|14:6|6|
|9:11|11|
|10:12|12|
|9:15|[Output Value]|

Return the final result as JSON in the format {"output_value": "<output value transformed from the input value>"}.
[A]:

(*\textbf{\underline{Completion:}}*)
{"output_value": "15"}
\end{lstlisting}
\begin{lstlisting}[title=Entity Matching Zero-Shot]
(*\textbf{\underline{Prompt:}}*)
Instruction: Please determine whether the First Cosmetic and the Second Cosmetic refer to the same entity or not. Your final answer should be 'Yes' or 'No'. Return the final result as JSON in the format {"answer": "<Yes or No>"}.  Let's think step by step and show your reasoning before showing the final result.

Question:
the First Cosmetic:
|Price|Color|Description|
|---|---|---|
|$16.50|Wheat |CLINIQUE brow shaper|

the Second Cosmetic can be presented as follows.
|Color|Description|Price|
|---|---|---|
|Charcoaled |Clinique Brow Shaper Charcoaled|$29.90|

Let's think step by step and show your reasoning before showing the final result. Return the final result as JSON in the format {"answer": "<Yes or No>"}.
Answer:

(*\textbf{\underline{Completion:}}*)
the First Cosmetic and the Second Cosmetic are different entities because they have different prices and colors. the First Cosmetic is priced at $16.50 and is Wheat in color, while the Second Cosmetic is priced at $29.90 and is Charcoaled in color. Therefore, the final answer is {"answer": "No"}.
\end{lstlisting}
\begin{lstlisting}[title=Entity Matching Few-Shot]
(*\textbf{\underline{Prompt:}}*)
Instruction: Please verify whether Cosmetic I and Cosmetic II are denoting the same entity. Indicate your conclusive answer as either 'Yes' or 'No'. Return the final result as JSON in the format {"answer": "<Yes or No>"}.  Let's think step by step and show your reasoning before showing the final result.

In:
Here is the description of Cosmetic I.
|Description|Price|Color|
|---|---|---|
|Hourglass extreme sheen high shine lip gloss|$28.00|Child|

Cosmetic II:
|Price|Color|Description|
|---|---|---|
|$41.35|Truth|Hourglass Extreme Sheen High Shine Lip Gloss Truth|

Out:
Cosmetic I and Cosmetic II are different entities because they have different prices ($28.00 vs. $41.35) and colors (Child vs. Truth). Therefore, the final answer is {"answer": "No"}.

In:
Here is the description of Cosmetic I.
|Description|Price|Color|
|---|---|---|
|tarte lipsurgence lip tint|$24.00|Grateful|

Cosmetic II:
|Price|Color|Description|
|---|---|---|
|$35.40|Energy|Tarte Lipsurgence Skintuitive Lip Tint - Energy (.10 oz)|

Out:
Cosmetic I and Cosmetic II are different entities because they have different prices, colors, and descriptions. Cosmetic I is a Tarte Lipsurgence Lip Tint with a price of $24.00 and a color of Grateful. Cosmetic II is a Tarte Lipsurgence Skintuitive Lip Tint with a price of $35.40 and a color of Energy. Therefore, the final answer is {"answer": "No"}.

In:
Here is the description of Cosmetic I.
|Description|Price|Color|
|---|---|---|
|Bobbi Brown foundation brush|$42.00|nan|

Cosmetic II:
|Price|Color|Description|
|---|---|---|
|$69.00|nan|Bobbi Brown Foundation Brush|

Out:
Cosmetic I and Cosmetic II refer to the same entity as they both have the same description, "Bobbi Brown Foundation Brush", and the same color, nan. The only difference is the price, with Cosmetic I being $42.00 and Cosmetic II being $69.00. Therefore, the final answer is {"answer": "Yes"}.

In:
Here is the description of Cosmetic I.
|Description|Price|Color|
|---|---|---|
|Dior glow maximizer primer|$42.00|1|

Cosmetic II:
|Price|Color|Description|
|---|---|---|
|$43.91|1|Dior Glow Maximizer Light Boosting Primer|

Out:
Cosmetic I and Cosmetic II are the same product, Dior Glow Maximizer Light Boosting Primer. Both have the same price ($42.00 and $43.91) and color (1). The only difference is the order of the columns. Therefore, the final answer is {"answer": "Yes"}.

In:
Here is the description of Cosmetic I.
|Description|Price|Color|
|---|---|---|
|Bobbi Brown foundation stick|$44.00|Warm Honey|

Cosmetic II:
|Price|Color|Description|
|---|---|---|
|$58.50|Warm Ivory|Exclusive Make Up Product By Bobbi Brown Foundation Stick - #1 Warm Ivory 9g/0.31oz|

Let's think step by step and show your reasoning before showing the final result. Return the final result as JSON in the format {"answer": "<Yes or No>"}.
Out:

(*\textbf{\underline{Completion:}}*)
Cosmetic I and Cosmetic II are different entities because they have different prices, colors, and descriptions. Cosmetic I is a Bobbi Brown foundation stick that costs $44.00 and is in the color Warm Honey. Cosmetic II is an Exclusive Make Up Product By Bobbi Brown Foundation Stick that costs $58.50 and is in the color Warm Ivory. Therefore, the final answer is {"answer": "No"}.
\end{lstlisting}
\begin{lstlisting}[title=Data Imputation Zero-Shot]
(*\textbf{\underline{Prompt:}}*)
Task: Please fill in the missing value in the input table. The missing value is denoted by '[MISSING]'. Please only return the value filled in. Do not return the whole table. Return the final result as JSON in the format {"value": "<value filled in>"}.

In:
|Title|Credited artist(s)|Year|
|---|---|---|
|"10 or a 2-Way"|Korn|2005|
|"4 U"|Korn|1999|
|"A Different World"|Korn Corey Taylor|2016|
|"A.D.I.D.A.S."|[MISSING]|1996|
|"Alive"|Korn|2003|

Return the final result as JSON in the format {"value": "<value filled in>"}.
Out:

(*\textbf{\underline{Completion:}}*)
{"value": "Korn"}
\end{lstlisting}
\begin{lstlisting}[title=Data Imputation Few-Shot]
(*\textbf{\underline{Prompt:}}*)
Objective: To fulfill your responsibility, please provide the missing value in the input table, represented by '[MISSING]'. Only the filled-in value should be provided, not the whole table. Return the final result as JSON in the format {"value": "<value filled in>"}.

Input:
|Column1|area|id_ov|
|---|---|---|
|16|CRM|29|
|25|CCS|15|
|32|CRM|4|
|12|CRM|9|
|8|[MISSING]|9|

Output:
{"value": "CRM"}

Input:
|Column1|area|id_ov|
|---|---|---|
|0|CRM|5|
|1|[MISSING]|9|
|2|CCS|9|
|3|CCS|9|
|4|CRM|9|

Return the final result as JSON in the format {"value": "<value filled in>"}.
Output:

(*\textbf{\underline{Completion:}}*)
{"value": "CCS"}
\end{lstlisting}
\begin{lstlisting}[title=Row/Column Sorting Zero-Shot]
(*\textbf{\underline{Prompt:}}*)
Description: Please sort the table by the column "Gauge" such that the values in this column are in alphabetically ascending order. Please return the sorted table.

Q:
|Gauge|Dia.|Delta|
|---|---|---|
|5/0|0.5|nan|
|5/0|0.5|0.046|
|4/0|0.454|0.046|
|4/0|0.454|0.029|
|3/0|0.425|0.029|

A:

(*\textbf{\underline{Completion:}}*)
|Gauge|Dia.|Delta|
|---|---|---|
|3/0|0.425|0.029|
|4/0|0.454|0.046|
|4/0|0.454|0.029|
|5/0|0.5|nan|
|5/0|0.5|0.046|
\end{lstlisting}
\begin{lstlisting}[title=Row/Column Sorting Few-Shot]
(*\textbf{\underline{Prompt:}}*)
Instruction: Could you sort the table by arranging it based on column headers in alphabetical descending order from left to right? After the sorting, kindly return the table.

Input:
|Rank|State|%|
|---|---|---|
|46.0|NH|18.7|
|45.0|MT|19.0|
|49.0|NY|18.3|
|48.0|OR|18.5|
|51.0|TX|18.1|

Output:
|State|Rank|%|
|---|---|---|
|NH|46.0|18.7|
|MT|45.0|19.0|
|NY|49.0|18.3|
|OR|48.0|18.5|
|TX|51.0|18.1|

Input:
|Rank|State|%|
|---|---|---|
|40.0|KS|20.0|
|41.0|GA|20.0|
|42.0|ND|19.6|
|43.0|VA|19.3|
|44.0|RI|19.3|

Output:

(*\textbf{\underline{Completion:}}*)
|State|Rank|%|
|---|---|---|
|KS|40.0|20.0|
|GA|41.0|20.0|
|ND|42.0|19.6|
|VA|43.0|19.3|
|RI|44.0|19.3|
\end{lstlisting}